\documentclass[lettersize, journal]{IEEEtran}
\usepackage{cite}
\usepackage{amsmath,amssymb,amsfonts}
\usepackage{algorithmic}
\usepackage{graphicx}
\usepackage{textcomp}
\usepackage{xcolor}
\usepackage{algorithm}
\usepackage{colortbl}
\usepackage{arydshln}

\usepackage{amsmath,amssymb}
\usepackage{caption}
\usepackage{subcaption}
\usepackage{url}
\usepackage{hyperref}
\usepackage{multirow}
\usepackage{bm}
\usepackage{overpic}
\usepackage{tikz}
\usepackage{booktabs}

\usepackage{hyperref}

\newcommand{\tabincell}[2]{\begin{tabular}{@{}#1@{}}#2\end{tabular}}
\begin{document}
\setcounter{secnumdepth}{4} %
\setcounter{tocdepth}{4} %

\title{Static for Dynamic: Towards a Deeper Understanding of Dynamic Facial Expressions Using Static Expression Data}
\author{Yin~Chen, Jia~Li, Yu~Zhang, Zhenzhen~Hu, Shiguang~Shan,~\IEEEmembership{Fellow,~IEEE,}

	Meng~Wang,~\IEEEmembership{Fellow,~IEEE,} and
	Richang~Hong,~\IEEEmembership{Member,~IEEE}

	\thanks{
		This work was supported by the National Key Research and Development Program of China under No. 2023YFC2506803. It was also supported in part by the National Natural Science Foundation of China (NSFC, No. 62202139 and No. 62172138), and in part by the Fundamental Research Funds for the Central Universities (No. JZ2025HGTB0226 and No. JZ2024HGTG0310). The computations were completed on the High-Performance Computing Platform of Hefei University of Technology.e. (Yin Chen and Jia Li contributed equally to this work.) (Corresponding author: Jia Li.)

		Yin Chen, Jia Li, Yu Zhang, Zhenzhen Hu, Meng Wang and Richang Hong are with the School of Computer
		Science and Information Engineering, Hefei University of Technology, Hefei
		230601, China (e-mail: chenyin@mail.hfut.edu.cn; jiali@hfut.edu.cn; yuz@mail.hfut.edu.cn; huzhen.ice@gmail.com;
		eric.mengwang@gmail.com; hongrc.hfut@gmail.com).

		Shiguang Shan is with the Key Laboratory of Intelligent Information Processing,
		Institute of Computing Technology, Chinese Academy of Sciences, Beijing 100190,
		China, and also with the University of Chinese Academy of Sciences, Beijing,
		100049, China (e-mail: sgshan@ict.ac.cn).

		© 2024 IEEE. Personal use of this material is permitted. Permission from IEEE must be obtained for all other uses, in any current or future media, including reprinting/republishing this material for advertising or promotional purposes, creating new collective works, for resale or redistribution to servers or lists, or reuse of any copyrighted component of this work in other works.
		Digital Object Identifier: \href{doi.org/10.1109/TAFFC.2025.3623135}{10.1109/TAFFC.2025.3623135}
		
	}
}

\maketitle

\begin{abstract}
	Dynamic facial expression recognition (DFER) infers emotions from the temporal evolution of expressions, unlike static facial expression recognition (SFER), which relies solely on a single snapshot. This temporal analysis provides richer information and promises greater recognition capability.
	However, current DFER methods often exhibit unsatisfied performance largely due to fewer training samples compared to SFER. Given the inherent correlation between static and dynamic expressions, we hypothesize that leveraging the abundant SFER data can enhance DFER.
	To this end, we propose Static-for-Dynamic (S4D), a unified dual-modal learning framework that integrates SFER data as a complementary resource for DFER.
	Specifically, S4D employs dual-modal self-supervised pre-training on facial images and videos using a shared Vision Transformer (ViT) encoder-decoder architecture, yielding improved spatiotemporal representations. The pre-trained encoder is then fine-tuned on static and dynamic expression datasets in a multi-task learning setup to facilitate emotional information interaction.
	Unfortunately, vanilla multi-task learning in our study results in negative transfer. To address this, we propose an innovative Mixture of Adapter Experts (MoAE) module that facilitates task-specific knowledge acquisition while effectively extracting shared knowledge from both static and dynamic expression data.
	Extensive experiments demonstrate that S4D achieves a deeper understanding of DFER, setting new state-of-the-art performance on FERV39K, MAFW, and DFEW benchmarks, with weighted average recall (WAR) of 53.65\%, 58.44\%, and 76.68\%, respectively. Additionally, a systematic correlation analysis between SFER and DFER tasks is presented, which further elucidates the potential benefits of leveraging SFER.
\end{abstract}


\section{Introduction}
\IEEEPARstart{F}{acial} expression recognition (FER) is essential in fields such as human-computer interaction~\cite{Liu2017AFE}, mental health diagnosis~\cite{Bisogni2022ImpactOD}, and driving safety~\cite{Wilhelm2019TowardsFE}. Traditional FER methods focus on static facial expression recognition (SFER), which captures single moments of expressions from images. However, static images cannot fully reflect the dynamic changes of emotions over time. In contrast, video-based data provides richer temporal information and offers a more comprehensive view of emotions, prompting a shift towards dynamic facial expression recognition (DFER)~\cite{9039580}.

\begin{figure}[t]
	\centering
	\begin{tikzpicture}
		\node[anchor=south west,inner sep=0] (image) at (0,0) {\includegraphics[width=0.9\linewidth]{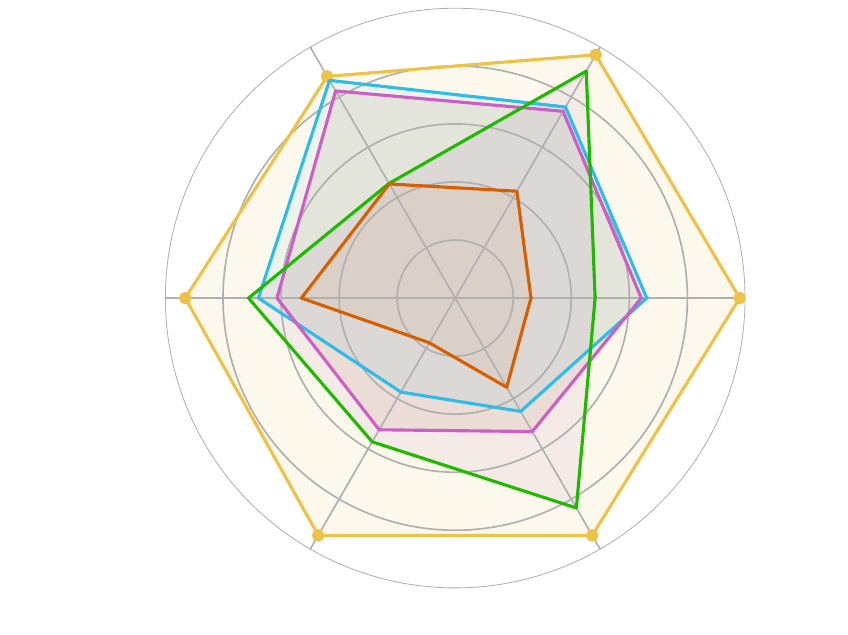}};
		\draw[dashed, color=gray] (4.29,3.16)-- (1.68,1.4) ;
		\draw[dashed, color=gray] (4.29,3.16)-- (6.9,1.4) ;
		\draw[dashed, color=gray] (4.29,3.16)-- (4.28,6.3) ;

		\node at (4.2,0.2) {MAFW};
		\node at (1.0,4.8) {FERV39K};
		\node at (7.3,4.8) {DFEW};
	\end{tikzpicture}
	\caption{Performance comparison between previous SOTA methods \cite{zhao2023prompting,sun2023mae,chen2023static,tong2022videomae} and our proposed S4D on FERV39K \cite{ferv39k2022}, MAFW \cite{MAFW}, and DFEW \cite{DFEW} datasets. Unweighted average recall (UAR, \%) and weighted average recall (WAR, \%) are reported.
		S4D,
		which incorporates static expression knowledge through a unified dual-modal learning framework,
		consistently outperforms the baseline method, VideoMAE \cite{tong2022videomae}, previously pre-trained on VoxCeleb2 \cite{chung2018voxceleb2}, across all these real-world DFER datasets.}
	\label{fig:compare_with_baseline}
\end{figure}

To support the development and evaluation of DFER algorithms, researchers have constructed various datasets, including lab-controlled datasets and in-the-wild datasets. Lab-controlled datasets, such as CK+ \cite{CK+5543262}, MMI \cite{MMI1521424}, and Oulu-CASIA \cite{4761697}, are collected under controlled laboratory environments and contain exaggerated expressions performed by participants under specific instructions. However, the applicability of these datasets in real-world scenarios is limited. To overcome this limitation, researchers have started building large-scale in-the-wild datasets, such as DFEW \cite{DFEW}, FERV39K \cite{ferv39k2022}, and MAFW  \cite{MAFW}. These datasets are collected from real-world scenarios, including movies, TV shows, and online videos, and encompass a wide range of head movements, illumination variations, and spontaneous expressions.
The emergence of these in-the-wild datasets has provided valuable data supporting the development of robust and practical DFER algorithms.

Despite the richer temporal cues inherent in video clips, which theoretically enable superior recognition, in-the-wild DFER lags behind its SFER counterpart. This performance disparity stems primarily from the limited scale and diversity of DFER datasets.
While SFER benefits from large, diverse datasets containing millions of images and labels, DFER datasets are relatively smaller and less varied \cite{chen2023static}. However, the intrinsic link between these tasks presents a valuable opportunity to improve DFER performance by leveraging SFER data. Specifically, SFER images often capture moments of pronounced emotional intensities and inherently contain highly discriminative features crucial for understanding dynamic expressions.  From a temporal perspective, these static images can be viewed as critical snapshots that capture essential moments within dynamic expression sequences.
Furthermore, the shared categorical labels and common semantic space between SFER and DFER, as evidenced by the semantic similarity of their class representations (see Fig. \ref{fig:similarity}), suggest that leveraging the extensive SFER data can significantly enhance DFER model training while reducing reliance on scarce, temporally annotated video data.

Motivated by the inherent connection between SFER and DFER, recent research has begun exploring various approaches to transfer static knowledge to the dynamic domain. Representative works include AEN \cite{Lee_2023_CVPR} and S2D \cite{chen2023static}. AEN~\cite{Lee_2023_CVPR} pioneered this direction by integrating multi-level semantic features extracted from SFER models and employing emotion-guided loss functions to enhance recognition accuracy. Subsequently, S2D~\cite{chen2023static} further advances the field by incorporating Temporal Modeling Adapters (TMAs) to extend pre-trained SFER models for DFER tasks.
Although these approaches have shown promising results, they primarily focus on feature-level knowledge transfer using pre-trained SFER models, potentially limiting the full exploitation of static expression information. \textbf{\textit{We argue that a more comprehensive approach, utilizing both data (i.e., images) and corresponding labels from static datasets, could better leverage SFER knowledge to enhance DFER performance}}.

Based on these observations, we propose a novel framework, Static-for-Dynamic (S4D),  which enhances the understanding of dynamic facial expressions by fully leveraging static expression data at both the input- and label- levels. To the best of our knowledge, we represent the first systematic exploration and analysis of the inherent correlations between SFER and DFER tasks, integrating both modalities in terms of input data (i.e., facial images and videos) and emotion labels. The S4D framework employs a unified dual-modal learning approach to seamlessly merge SFER and DFER tasks, thereby advancing DFER performance. Specifically, S4D comprises two key stages: Dual-Modal Pre-Training and Joint Fine-Tuning. During the pre-training stage, we utilize a shared Vision Transformer (ViT) \cite{dosovitskiy2020image} encoder-decoder architecture combined with Masked Autoencoders (MAE) \cite{he2022masked} to perform masked modeling on both image and video modalities. This approach allows the model to efficiently learn generalizable spatiotemporal representations from both static and dynamic facial data. In the fine-tuning stage, the pre-trained encoder is jointly optimized using both FER image and video datasets with a multi-task learning setup, facilitating cross-modal interaction of emotional information. However, as highlighted in previous works \cite{kollias2023multi, kollias2024distribution}, directly employing multi-task learning in related FER tasks may result in a negative transfer. We also observed similar performance degradation when applying straightforward multi-task learning to SFER and DFER tasks. To address this issue, we introduce a novel Mixture of Adapter Experts (MoAE) module, which operates in parallel with the original Feed-Forward Network (FFN) in the ViT layers. This innovative design enables the model to simultaneously capture task-agnostic knowledge through the FFN and task-specific features via the MoAE, thereby preventing negative transfer and enhancing emotional information interaction. By integrating the MoAE module within the unified dual-modal learning framework, the S4D framework effectively integrates complementary information from static facial expression data, leading to a deeper understanding of dynamic facial expressions and significant improvements in DFER performance.
The code and models are publicly available at \url{https://github.com/MSA-LMC/S4D}.

We summarize our main contributions as follows:
\begin{itemize}
	\item[$\bullet$]
		\textbf{S4D, a Unified Dual-Modal Learning Framework.}
		This framework seamlessly integrates dual-modal pre-training and joint fine-tuning on FER images and videos. Such a comprehensive approach yields rich and powerful spatiotemporal representations during pre-training and significantly improves expression semantic understanding through joint fine-tuning, ultimately achieving superior DFER performance. Incidentally, our final single model can perform both SFER and DFER tasks simultaneously.

	\item[$\bullet$]
		\textbf{MoAE, a Mixture of Adapter Experts Module.}
		We incorporate the MoAE module into the ViT layers of the S4D encoder, operating in parallel with the FFN. This design allows FFN to focus on task-agnostic knowledge, while MoAE captures task-specific knowledge, effectively alleviating the negative transfer between SFER and DFER tasks and enabling more discriminative feature learning.

	\item[$\bullet$]
		\textbf{Analysis of Task Correlation and SOTA Performance.}
		We present a systematic correlation analysis between SFER and DFER tasks from the perspectives of semantic and expert pathways, providing valuable insights into their inherent characteristics.
		Additionally, as shown in Fig. \ref{fig:compare_with_baseline}, our approach achieves substantial improvements over existing state-of-the-art methods in unweighted average recall (UAR) and weighted average recall (WAR), particularly surpassing the VideoMAE \cite{tong2022videomae} baseline.

\end{itemize}

\section{Related Work}

\begin{figure*}[!t]
	\centering
	\includegraphics[width=1.0\linewidth]{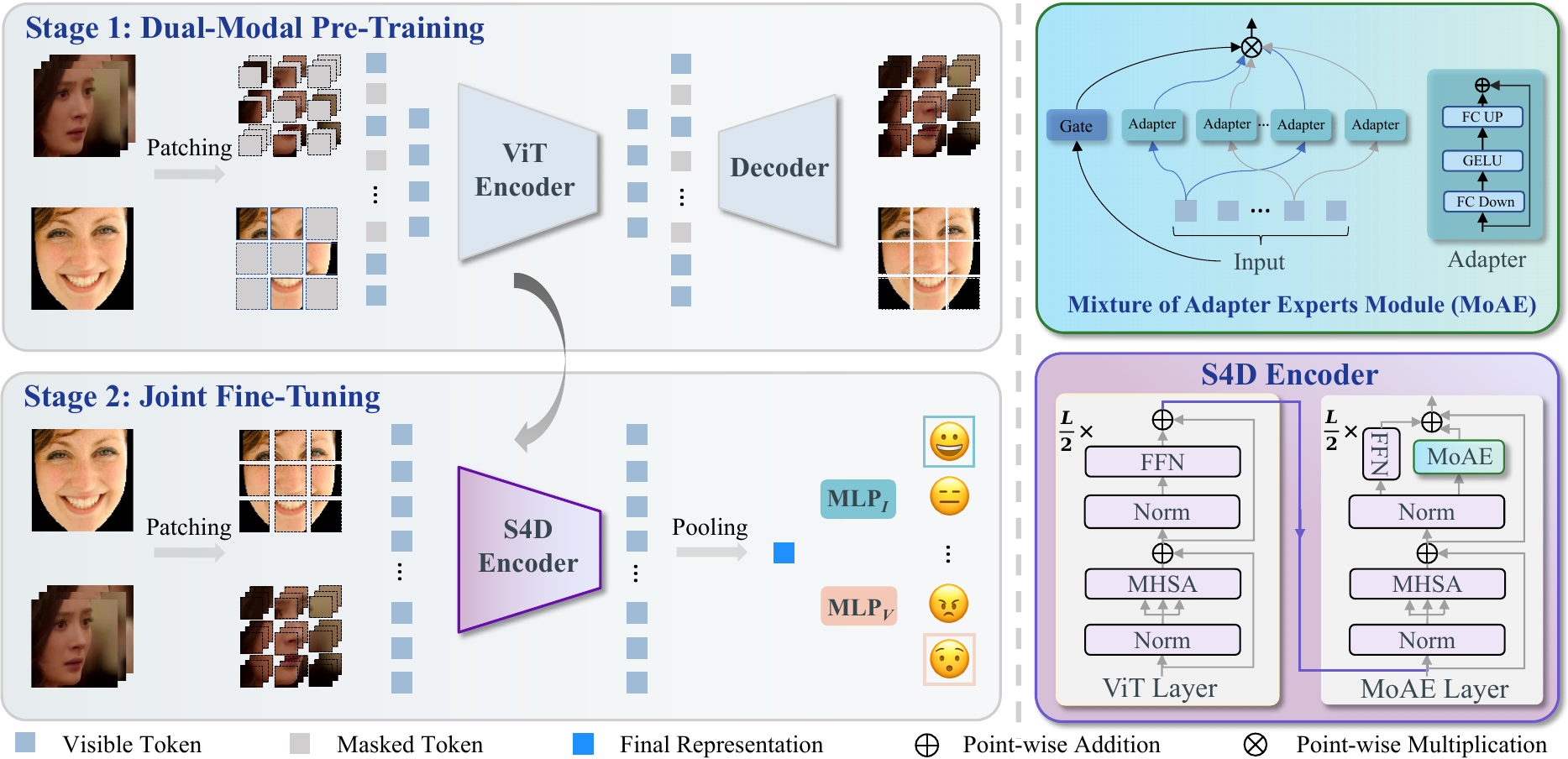}
	\caption{\textbf{Overview of our proposed S4D framework.} We utilize Vision Transformer (ViT) \cite{dosovitskiy2020image}  as the backbone and pre-train it on facial image and video datasets using Masked Autoencoders \cite{feichtenhofer2022masked}. The pre-trained ViT encoder is then used to initialize the S4D encoder, which is further fine-tuned on static and dynamic FER datasets. The proposed Mixture of Adapter Experts (MoAE) module is integrated into the ViT layers to create MoAE layers during joint fine-tuning. MLP$_I$ and MLP$_V$ denote the classification heads for SFER and DFER, while FFN, Norm, and MHSA represent the feed-forward network, layer normalization, and multi-head self-attention mechanisms, respectively.}
	\label{fig:overview}
\end{figure*}

\subsection{Dynamic Facial Expression Recognition}

The evolution of Dynamic Facial Expression Recognition (DFER) has shifted from traditional handcrafted feature-based methods to advanced deep learning approaches for in-the-wild scenarios. Recent DFER approaches can be broadly categorized into four distinct groups. First, end-to-end supervised learning initially used 3D CNNs like C3D~\cite{tran2015learning}, R(2+1)D~\cite{tran2018closer}, and I3D-RGB~\cite{carreira2017quo} for spatiotemporal feature learning, later combining 2D CNNs~\cite{krizhevsky2012imagenet} with RNNs or LSTMs~\cite{graves2012long, wang2022dpcnet} for temporal modeling, and more recently adopting Transformer-based architectures~\cite{vaswani2017attention, ma2022spatio, zhao2021former}.
	Second, CLIPER~\cite{li2023cliper}, DFER-CLIP~\cite{zhao2023prompting} and A$^{3}$lign-DFER~\cite{align-dfer} utilize vision-language models, notably CLIP~\cite{radford2021learning} to improve DFER performance by incorporating semantic understanding.
	Thid, ~\cite{sun2023mae, sun2023svfap, sun2024hicmae} employ self-supervised learning to exploit large amounts of unlabeled facial video data. Notably, MAE-DFER~\cite{sun2023mae} utilizes a local-global interaction Transformer encoder for masked reconstruction, improving task-specific learning. Finally, recent works explore transfer learning from static knowledge to enhance DFER performance \cite{chen2023static, Lee_2023_CVPR}. For instance, AEN~\cite{Lee_2023_CVPR} combines multi-level semantic features from SFER models with emotion-guided loss functions, while S2D~\cite{chen2023static} adapts a pre-trained SFER model to DFER via Temporal Modeling Adapters.  Unlike these methods, which focus on single-modal or task-specific learning, our S4D framework integrates both static and dynamic facial expression data through a unified dual-modal learning approach, offering a more comprehensive and promising solution for DFER.
\vspace{-5pt} %
\subsection{Multi-Modal Learning and Unified Modeling}

Multi-modal learning has significantly advanced computer vision by integrating diverse data modalities, such as image, text, video, and audio~\cite{gong2014improving, lu202012, morgado2021robust, morgado2021audio}. These techniques have also been applied to FER tasks~\cite{li2023cliper, align-dfer, sun2024hicmae}. Traditional methods often train different modalities independently before aligning them, which may overlook the potential benefits of a unified multi-modal learning approach. Recent advancements have aimed to address this issue by unifying the learning process~\cite{hu2021unit, girdhar2023omnimae}.  For instance, BEVT~\cite{wang2022bevt} and OmniMAE~\cite{girdhar2023omnimae} utilize a single encoder to handle both image and video modalities during pre-training while CoVER~\cite{zhang2021co} employs a unified model for multiple visual datasets and tasks during fine-tuning. However, these methods often fail to maintain learning consistency between the pre-training and fine-tuning stages. In contrast,  S4D introduces a unified learning framework integrating facial images and video processing, ensuring consistent multi-modal learning across stages. This novel approach not only overcomes the limitations of previous methods but also significantly improves DFER performance.

\section{Methodology}
\subsection{Overview}
Fig. \ref{fig:overview} illustrates the architecture of the S4D framework, which consists of two main stages: Dual-Modal Pre-Training and Joint Fine-Tuning. In the pre-training phase, the framework utilizes the Vision Transformer (ViT)~\cite{dosovitskiy2020image} with Masked Autoencoder (MAE)~\cite{he2022masked} to reconstruct both facial images and videos, learning powerful spatiotemporal representations. After pre-training, the ViT encoder is jointly fine-tuned on both static and dynamic FER datasets in a multi-task learning setup. The final representations, obtained from either images or videos, are fed into separate classification heads for SFER and DFER tasks, with the network jointly optimized via cross-entropy loss. To address the potential risk of negative transfer during direct multi-task joint fine-tuning, the Mixture of Adapter Experts (MoAE) modules are introduced in the deeper layers of ViT to enable generic and task-specific knowledge acquisition.
The Dual-Modal Pre-Training, Joint Fine-Tuning, and the MoAE module will be detailed in the following Section \ref{sec:pretrain}, Section \ref{sec:finetune}, and Section \ref{sec:moae}, respectively.

\subsection{Stage 1: Dual-Modal Self-Supervised Pre-Training}
\label{sec:pretrain}
During the pre-training phase, we employ the MAE \cite{feichtenhofer2022masked}  strategy to jointly train a standard ViT model on large-scale facial image and video datasets. MAE is a self-supervised learning approach that randomly masks a portion of the input data and trains the model to reconstruct the original data from the masked input. This encourages the model to learn
powerful representations by capturing the underlying structure and semantics of
the data.

Following OmniMAE \cite{girdhar2023omnimae}, we treat both image and video inputs as 4D tensors $\bm{X} \in \mathbb{R}^{T \times H \times W \times C}$, where $T$, $H$, $W$, and $C$ represent the number of frames, height, width, and channels, respectively. In this context, we consider an image as a special case of a video with a single frame, thus setting $T=1$. Given an input tensor $\bm{X}$ sampled from image or video datasets, we first generate a random binary mask tensor $\bm{M}^{T \times H \times W} \in \{0, 1\}$ with a predefined mask ratio to decide where to drop the patches. The masked input tensor $\bm{X}_{m}$ is computed as
the element-wise product of $\bm{X}$ and $\bm{M}$:
\begin{equation}
	\bm{X}_{m} = \bm{X} \odot \bm{M}.
\end{equation}

$\bm{X}_{m}$ is then fed into the ViT encoder $f_{E}$ to
obtain the latent representation $\bm{Z} = f_{E}(\bm{X}_{m})$. Subsequently,
the decoder $f_{D}$ takes $\bm{Z}$ as input and attempts to reconstruct the
masked pixels of $\bm{X}$. The objective of the pre-training phase is
to minimize the reconstruction loss, which is defined as the mean squared error
(MSE) between the reconstructed tensor $\hat{\bm{X}} = f_{D}(\bm{Z})$ and the
unmasked tensor $\bm{X}$:
\begin{equation}
	\mathcal{L}_{\mathrm{MASK}} = \frac{1}{ \sum \bm{M} } \sum (1-\bm{M}) \odot (\bm{\hat{X}} - \bm{X})^2.
\end{equation}

\subsection{Stage 2: Joint Fine-Tuning on Static and Dynamic Data}
\label{sec:finetune}

We propose a joint fine-tuning strategy utilizing both SFER and DFER datasets to maintain consistency with dual-modal pre-training and fully exploit knowledge from both domains. This approach mines complementary spatial information from SFER data and temporal dynamics from DFER data, enabling the model to learn more robust, generalizable representations. By integrating these two information sources, the model gains a deeper and more comprehensive understanding of dynamic facial expressions.

As illustrated in Fig. \ref{fig:overview}, the pre-trained ViT encoder, $f_{E}$, is employed to initialize our S4D encoder, $f_U$.  The S4D encoder is then jointly fine-tuned on SFER and DFER tasks using the provided inputs ${(\bm{X}_i, y_i)}$, where $\bm{X}_i$ represents the visual inputs and $y_i$ denotes the corresponding labels. During the fine-tuning process, $f_U$ generates a unified embedding $\bm{\Phi} =f_{U}(\bm{X})$ for both image and video inputs. The final prediction for each task is generated by a separate task-specific Multi-Layer Perceptron (MLP$_I$ for SFER and MLP$_V$ for DFER) applied to the final representation $\bm{\Phi}$. To optimize the model, we minimize the cross-entropy loss on the training datasets using mini-batch stochastic gradient descent. Each mini-batch is constructed independently from the SFER or DFER datasets. This approach employs single-source mini-batch sampling, maximizing GPU efficiency by eliminating the need to pad token sequences due to differences in the number of patches between images and videos.

The total loss for joint fine-tuning is defined as follows:
\begin{equation}
	\mathcal{L}_{\mathrm{total}}=(1-\alpha) \cdot \mathcal{L}_{\mathrm{SFER}}+\alpha \cdot \mathcal{L}_{\mathrm{DFER}},
\end{equation}
where $\mathcal{L}_{\mathrm{SFER}}$ and $\mathcal{L}_{\mathrm{DFER}}$ are the cross-entropy losses for SFER and DFER tasks, respectively. The binary indicator $\alpha \in \{0, 1\}$ toggles between SFER ($\alpha=0$) and DFER ($\alpha=1$) tasks based on the mini-batch source.

\subsection{Mixture of Adapter Experts Module}
\label{sec:moae}
During the fine-tuning stage, we incorporate static expression data alongside DFER datasets to improve the model’s performance. However, using simple multi-task learning approaches, such as multiple classification heads, may lead to negative transfer, failing to fully exploit the correlation between SFER and DFER, thus hindering optimal performance.
To address this issue and fully utilize the correlations between different SFER and DFER tasks, we propose the Mixture of Adapter Experts (MoAE) module.

The design of the MoAE module is inspired by the Mixture of Experts (MoE) \cite{jacobs1991adaptive}, which employs multiple expert networks
coordinated by a gating mechanism. In the MoE framework, a gating network $G$ is responsible for assigning weights to $n$ independent experts
$\{E_i\}^n_{i=1}$ based on the input $\bm{x}$. The gating network computes a weight distribution by applying a $\operatorname{Softmax}$ function to the dot
product of the input  and a learnable matrix $\bm{W}_g$:
\begin{equation}
	G(\bm{x}) = \operatorname{Softmax}(\bm{x} \cdot {\bm{W}_g}).
\end{equation}

To encourage load balance, we adopt Noisy Top-K Gating \cite{shazeer2017outrageously}. This gating mechanism
introduces noise to the logits before applying the $\operatorname{Top-K}$
operation and $\operatorname{Softmax}$ function:
\begin{equation}
	H(\bm{x}) = \bm{x} \cdot \bm{W}_g + \epsilon,
\end{equation}
\begin{equation}
	G(\bm{x}) = \operatorname{Softmax}(\operatorname{Top-K}(H(\bm{x}), k)),
	\label{eq:topk}
\end{equation}
where $\epsilon \sim \mathcal{N}(0, \sigma^{2} \boldsymbol{I})$ is Gaussian noise with mean 0 and variance $\sigma^{2}$, and $\operatorname{Top-K}(H(\bm{x}), k)$ retains the top $k$ largest values from $H(\bm{x})$ while suppressing the remaining elements to negative infinity. The introduction of noise helps to diversify the expert selection, preventing the gating network from always choosing the same experts. The output of the MoE layer is then computed as a weighted sum of the expert outputs:
\begin{equation}
	\operatorname{MoE}(\bm{x}) = \sum_{i=1}^{n} G(\bm{x})_i \cdot E_i(\bm{x}).
\end{equation}

Previous works, such as ViT-MoE \cite{NEURIPS2021_48237d9f}, often use Feed-Forward Networks (FFNs) as experts in the MoE, significantly increasing the model parameters but potentially compromising structure and performance.
To preserve the model structure and learn task-specific knowledge with minimal parameter increase, we employ parameter-efficient adapters \cite{houlsby2019parameter} as experts in MoAE. These adapters are lightweight modules consisting of two
linear layers with a $\operatorname{GELU}$ activation function:
\begin{equation}
	\operatorname{Adapter}(\bm{x}) = \bm{x} + \bm{W}_2(\operatorname{GELU}(\bm{W}_1 \bm{x} + \bm{b}_1)) + \bm{b}_2,
\end{equation}
where $\bm{W}_1 \in \mathbb{R}^{d \times r}, \bm{W}_2 \in \mathbb{R}^{r \times d}, \bm{b}_1 \in \mathbb{R}^{r}, \bm{b}_2 \in \mathbb{R}^{d}$ are learnable parameters. The
input/output dimension $d$ and bottleneck dimension $r$ (set to $d/4$ in our work) satisfy $r \ll d$, allowing the adapter to learn representations with minimal additional parameters. The formulation of the MoAE module can be expressed as:
\begin{equation}
	\operatorname{MoAE}(\bm{x}) = \sum_{i=1}^{n} G(\bm{x})_i \cdot \operatorname{Adapter}_i(\bm{x}),
\end{equation}
where $\operatorname{Adapter}_i$ denotes the $i$-th adapter expert and $G(\bm{x})$ is the gating network determining each adapter expert's contribution based on input $\bm{x}$.

Unlike ViT-MoE, which replaces the FFN with MoE, we integrate the MoAE module into the ViT layer, resulting in the MoAE layer, which operates in parallel with the original FFN. This design, as shown in Fig. \ref{fig:overview}, replaces the latter half of the ViT layers with MoAE layers, preserving the original structure while enabling the FFN to focus on task-agnostic knowledge and the MoAE to capture task-specific insights. Additionally, our approach differs from the Mixture of Parameter-Efficient Experts method \cite{cai2024survey} commonly used in large language models by mitigating negative transfer and enhancing adaptability.
The computation flow in the MoAE layer is as follows:
\begin{equation}
	\bm{x}' = \bm{x} + \operatorname{MHSA}(\operatorname{LayerNorm}(\bm{x})),
\end{equation}
where $\operatorname{MHSA}$ is multi-head self-attention, $\operatorname{LayerNorm}$ is layer normalization, and $\bm{x}'$ represent the global relational representation. $\bm{x}'$ is then processed by FFN and MoAE:
\begin{equation}
	\bm{x}_g = \operatorname{FFN}(\operatorname{LayerNorm}(\bm{x}')),
\end{equation}
\begin{equation}
	\bm{x}_s = \operatorname{MoAE}(\operatorname{LayerNorm}(\bm{x}')),
\end{equation}
where $\bm{x}_g$ and $\bm{x}_s$  represent generic knowledge and task-specific knowledge, respectively. Finally, the output $\bm{x}_o$ of MoAE layer is calculated as:
\begin{equation}
	\bm{x}_o = \bm{x}' + \bm{x}_g + \bm{x}_s.
\end{equation}

\section{Experiments}
\subsection{Datasets}
\textbf{Pre-Training Datasets.} We conduct dual-modal self-supervised pre-training on two large-scale facial datasets: VoxCeleb2 \cite{chung2018voxceleb2} and AffectNet \cite{Mollahosseini2017AffectNetAD}. VoxCeleb2 is a comprehensive audio-visual speaker recognition dataset sourced from YouTube, comprising over 1 million utterances by 6,112 speakers. For pre-training, we utilize the development set, which includes 1,092,009 video clips extracted from 145,569 videos. AffectNet is a large-scale static facial expression dataset containing over 1 million images, of which 450,000 are manually annotated. AffectNet is divided into two subsets: AffectNet-8, featuring eight basic expressions, and AffectNet-7, featuring seven basic expressions. A total of 287,568 images from the AffectNet-8 training set are combined with the VoxCeleb2 dataset for the pre-training process.

\noindent \textbf{Fine-Tuning Datasets:} We evaluate our method on three widely recognized DFER benchmarks: DFEW~\cite{DFEW}, FERV39K~\cite{ferv39k2022}, and MAFW~\cite{MAFW}. The DFEW dataset comprises 16,372 video clips sourced from over 1,500 films, each annotated with one of seven basic expressions. FERV39K, the largest in-the-wild DFER dataset, includes 38,935 clips representing seven basic expressions across diverse scenarios, divided into 31,088 training clips and 7,847 testing clips. MAFW is a multimodal DFER dataset containing 10,045 video clips; however, our focus is solely on the video modality, which consists of 9,172 clips categorized into 11 classes. For the MAFW dataset, we employed MTCNN~\cite{zhang2016joint} to extract and align facial regions, effectively removing irrelevant backgrounds. Additionally, AffectNet-7~\cite{Mollahosseini2017AffectNetAD} was utilized for joint training alongside the DFER datasets. Table \ref{tab:datasets} summarizes the key characteristics of the above datasets.

\begin{table}[!t]
	\centering
	\setlength{\tabcolsep}{1mm} %

	\caption{A summary of the basic information about the pre-training and fine-tuning datasets used in this paper. V: video, I: image, CV: cross-validation.}

	\resizebox{\linewidth}{!}{
		\begin{tabular}{lcccc}
			\toprule

			Dataset                        & Samples   & Classes & Modality & Evaluation    \\
			\midrule

			\textbf{Pre-Training Datasets} &           &         &          &               \\
			VoxCeleb2 (dev)                & 1,092,009 & -       & V        & -             \\
			AffectNet-8 (train)            & 287,568   & -       & I        & -             \\
			\midrule

			\textbf{Fine-Tuning Datasets}  &           &         &          &               \\
			DFEW                           & 11,697    & 7       & V        & 5-fold CV     \\
			FERV39K                        & 38,935    & 7       & V        & Train \& test \\
			MAFW                           & 10,045    & 11      & V        & 5-fold CV     \\
			AffectNet-7                    & 287,401   & 7       & I        & Train \& test \\
			\bottomrule
		\end{tabular}
	}
	\label{tab:datasets}
\end{table}

We evaluate the performance using unweighted average recall (UAR) and weighted average recall (WAR), consistent with previous studies \cite{sun2023mae,chen2023static}. For DFEW and MAFW, we employ 5-fold cross-validation, aggregating predictions and labels across all folds for final UAR and WAR computation.

\begin{table*}[!t]
	\centering
	\caption{Comparisons of our S4D with the state-of-the-art DFER methods on DFEW, FERV39k, and MAFW. Baseline results are directly extracted from \cite{sun2023mae}. The best results are highlighted in bold, and the second-best underlined. TI:Time Interpolation; DS: Dynamic Sampling.}
	\setlength {\tabcolsep }{ 2mm }
	\resizebox{\linewidth}{!}{

		\begin{tabular}{lcccccccc}
			\toprule
			\multirow{2}{*}{Method}                    & \multirow{2}{*}{\tabincell{c}{Sample                                                                                                                                                                                                                                                                                 \\Strategies}     } & \multirow{2}{*}{Backbone} & \multicolumn{2}{c}{DFEW} & \multicolumn{2}{c}{FERV39k} & \multicolumn{2}{c}{MAFW}                                                             \\
			\cmidrule(lr){4-5} \cmidrule(lr){6-7} \cmidrule(lr){8-9}
			                                           &                                      &                   & UAR (\%)                                & WAR (\%)                                & UAR  (\%)                               & WAR (\%)                                & UAR (\%)                                & WAR (\%)                                \\
			\midrule

			EC-STFL ~\cite{DFEW}                       & TI                                   & C3D / P3D         & 45.35                                   & 56.51                                   & -                                       & -                                       & -                                       & -                                       \\
			Former-DFER ~\cite{zhao2021former}         & DS                                   & Transformer       & 53.69                                   & 65.70                                   & 37.20                                   & 46.85                                   & 31.16                                   & 43.27                                   \\
			CEFLNet ~\cite{liu2022clip}                & Clip-based                           & ResNet-18         & 51.14                                   & 65.35                                   & -                                       & -                                       & -                                       & -                                       \\
			NR-DFERNet ~\cite{li2022nr}                & DS                                   & CNN-Transformer   & 54.21                                   & 68.19                                   & 33.99                                   & 45.97                                   & -                                       & -                                       \\
			STT ~\cite{ma2022spatio}                   & DS                                   & ResNet-18         & 54.58                                   & 66.65                                   & 37.76                                   & 48.11                                   & -                                       & -                                       \\
			EST ~\cite{liu2023expression}              & DS                                   & ResNet-18         & 53.94                                   & 65.85                                   & -                                       & -                                       & -                                       & -                                       \\
			Freq-HD ~\cite{tao2023freq}                & FreqHD                               & VGG13-LSTM        & 46.85                                   & 55.68                                   & 33.07                                   & 45.26                                   & -                                       & -                                       \\
			LOGO-Former ~\cite{ma2023logo}             & DS                                   & ResNet-18         & 54.21                                   & 66.98                                   & 38.22                                   & 48.13                                   & -                                       & -                                       \\
			IAL ~\cite{li2023intensity}                & DS                                   & ResNet-18         & 55.71                                   & 69.24                                   & 35.82                                   & 48.54                                   & -                                       & -                                       \\
			AEN ~\cite{Lee_2023_CVPR}                  & DS                                   & ResNet-18         & 56.66                                   & 69.37                                   & 38.18                                   & 47.88                                   & -                                       & -                                       \\
			T-MEP~\cite{zhang2023transformer}          & DS                                   & Transformer       & 57.16                                   & 68.85                                   & -                                       & -                                       & 39.37                                   & 52.85                                   \\

			M3DFEL ~\cite{wang2023rethinking}          & DS                                   & ResNet-18-3D      & 56.10                                   & 69.25                                   & 35.94                                   & 47.67                                   & -                                       & -                                       \\
			CLIPER ~\cite{li2023cliper}                & DS                                   & CLIP-ViT-B/16     & 57.56                                   & 70.84                                   & 41.23                                   & 51.34                                   & -                                       & -                                       \\

			DFER-CLIP ~\cite{zhao2023prompting}        & DS                                   & CLIP-ViT-B/32     & 59.61                                   & 71.25                                   & 41.27                                   & 51.65                                   & 39.89                                   & 52.55                                   \\
			EmoCLIP ~\cite{foteinopoulou_emoclip_2024} & DS                                   & CLIP-ViT-B/32     & 58.04                                   & 62.12                                   & 31.41                                   & 36.18                                   & 34.24                                   & 41.46                                   \\

			A$^3$lign-DFER ~\cite{align-dfer}          & DS                                   & CLIP-ViT-L/14     & \underline{64.09}                       & 74.20                                   & 41.87                                   & 51.77                                   & 42.07                                   & 53.24                                   \\

			SVFAP ~\cite{sun2023svfap}                 & DS                                   & ViT-B/16          & 62.83                                   & 74.27                                   & 42.14                                   & 52.29                                   & 41.19                                   & 54.28                                   \\

			HiCMAE ~\cite{sun2024hicmae}               & DS                                   & ViT-B/16          & 63.76                                   & 75.01                                   & -                                       & -                                       & \underline{42.65}                       & 56.17                                   \\

			MAE-DFER ~\cite{sun2023mae}                & DS                                   & ViT-B/16          & 63.41                                   & 74.43                                   & 43.12                                   & 52.07                                   & 41.62                                   & 54.31                                   \\

			S2D ~\cite{chen2023static}                 & DS                                   & ViT-B/16          & 61.82                                   & \underline{76.03}                       & 41.28                                   & \underline{52.56}                       & 41.86                                   & \underline{57.37}                       \\

			\midrule
			\rowcolor{cyan!10}
			VideoMAE (Baseline) ~\cite{sun2023mae}     & DS                                   & ViT-B/16          & 63.60                                   & 74.60                                   & \underline{43.33}                       & 52.39                                   & 40.87                                   & 53.51                                   \\

			\rowcolor{cyan!10}
			\textbf{S4D (Ours)}                        & DS                                   & \textbf{ViT-B/16} & \textbf{66.80} (\textcolor{red}{+3.20}) & \textbf{76.68} (\textcolor{red}{+2.08}) & \textbf{43.41} (\textcolor{red}{+0.08}) & \textbf{53.65} (\textcolor{red}{+1.26}) & \textbf{43.72} (\textcolor{red}{+2.85}) & \textbf{58.44} (\textcolor{red}{+4.93}) \\		%
			\bottomrule
		\end{tabular}
	}
	\label{tab:sota}
\end{table*}

\subsection{Implementation Details}

We adopt ViT-B/16 \cite{dosovitskiy2020image} as the backbone for both our S4D, following VideoMAE \cite{tong2022videomae} and MAE-DFER \cite{sun2023mae}.  MAE-DFER initialized VideoMAE model with the weights pretrained exclusively on the VoxCeleb2 dataset and fine-tuned on the DFER dataset, serving as a fair and comparable baseline. To reduce background noise, we crop a central 160$\times$160 patch from each 224$\times$224 frame in the VoxCeleb2 dataset, following \cite{sun2023mae}. For pre-training, we sample 16 frames per clip with a temporal stride of 4, use a patch size of $2\times16\times16$. Static images are resized to 160$\times$160 and temporally replicated to match the patch size. We apply random masking at a ratio of 95\% for videos and 90\% for images, consistent with \cite{girdhar2023omnimae}. Training is performed using the AdamW optimizer with $\beta_1=0.9$ and $\beta_2=0.95$, an over batch size $N_\mathrm{bs}$ of 384, and a base learning rate $lr_{\mathrm{base}}$ of 1.6e-3, with weight decay set to 0.05. The learning rate is scaled linearly based on batch size: $lr=lr_{\mathrm{base}} \times \frac{N_\mathrm{bs}}{512}$. Pre-training is conducted on the VoxCeleb2 and AffectNet-8 datasets for 100 epochs, with a cosine learning rate scheduler.

For fine-tuning, we replace the latter half of the ViT layers with MoAE layers and initialize the model using the pre-trained ViT encoder weights. We employ a full fine-tuning strategy, where all parameters of the model are updated during training. All input frames and images are resized to \(160 \times 160\). The batch size is 32 for the DFER datasets and 64 for the SFER datasets, with a learning rate of \(4 \times 10^{-5}\). We sample 16 frames per clip with a temporal stride of 4 across all datasets, except for FERV39K, which uses a stride of 1 due to its unique properties. In the MoAE module, we set $k=2$ and the number of experts $n=8$. During joint fine-tuning, the proportion of the SFER dataset in each epoch is empirically set to 50\%. We optimize the model using AdamW for 100 epochs. During inference, following \cite{sun2023mae, chen2023static}, we uniformly sample two clips per video and averaging their predictions for DFER tasks. Unless otherwise specified, all results are obtained using the S4D (ViT+MoAE) model with the best-performing weights.

All pre-training experiments are performed on two Nvidia A800 GPUs, and fine-tuning experiments on two 4090 GPUs, utilizing the PyTorch framework.

\begin{table*}[!t]
	\centering
	\caption{Comparative analyses of accuracy across various emotion categories: S4D \textit{vs.} other approaches on DFEW. The best results are highlighted in bold, and the second-best underlined. Baseline results are directly extracted from  \cite{sun2023mae}.}
	\setlength {\tabcolsep }{ 1mm }
	\resizebox{\linewidth}{!}{
		\begin{tabular}{lccccccccc}
			\toprule
			\multirow{2}{*}{Method}                 & \multicolumn{7}{c}{Accuracy of Each Emotion} & \multicolumn{2}{c}{DFEW}                                                                                                                                                                                                                                                                                                                                   \\
			\cmidrule(lr){2-8} \cmidrule(lr){9-10}

			                                        & Happy                                        & Sad                                        & Neutral                                     & Angry                                      & Surprise                                & Disgust                                  & Fear                                     & UAR  (\%)                               & WAR   (\%)                               \\
			\midrule
			C3D  ~\cite{tran2015learning}           & 75.17                                        & 39.49                                      & 55.11                                       & 62.49                                      & 45.00                                   & 1.38                                     & 20.51                                    & 42.74                                   & 53.54                                    \\
			R(2+1)D-18  ~\cite{tran2018closer}      & 79.67                                        & 39.07                                      & 57.66                                       & 50.39                                      & 48.26                                   & 3.45                                     & 21.06                                    & 42.79                                   & 53.22                                    \\
			3D ResNet-18  ~\cite{he2016deep}        & 76.32                                        & 50.21                                      & 64.18                                       & 62.85                                      & 47.52                                   & 0.00                                     & 24.56                                    & 46.52                                   & 58.27                                    \\
			EC-STFL  ~\cite{DFEW}                   & 79.18                                        & 49.05                                      & 57.85                                       & 60.98                                      & 46.15                                   & 2.76                                     & 21.51                                    & 45.35                                   & 56.51                                    \\
			ResNet-18+LSTM  ~\cite{zhao2021former}  & 83.56                                        & 61.56                                      & 68.27                                       & 65.29                                      & 51.26                                   & 0.00                                     & 29.34                                    & 51.32                                   & 63.85                                    \\
			ResNet-18+GRU  ~\cite{zhao2021former}   & 82.87                                        & 63.83                                      & 65.06                                       & 68.51                                      & 52.00                                   & 0.86                                     & 30.14                                    & 51.68                                   & 64.02                                    \\

			Former-DFER  ~\cite{zhao2021former}     & 84.05                                        & 62.57                                      & 67.52                                       & 70.03                                      & 56.43                                   & 3.45                                     & 31.78                                    & 53.69                                   & 65.70                                    \\
			CEFLNet  \cite{liu2022clip}             & 84.00                                        & 68.00                                      & 67.00                                       & 70.00                                      & 52.00                                   & 0.00                                     & 17.00                                    & 51.14                                   & 65.35                                    \\
			NR-DFERNet  ~\cite{li2022nr}            & 88.47                                        & 64.84                                      & 70.03                                       & 75.09                                      & 61.60                                   & 0.00                                     & 19.43                                    & 54.21                                   & 68.19                                    \\
			STT  ~\cite{ma2022spatio}               & 87.36                                        & 67.90                                      & 64.97                                       & 71.24                                      & 53.10                                   & 3.49                                     & 34.04                                    & 54.58                                   & 66.65                                    \\
			EST  ~\cite{liu2023expression}          & 86.87                                        & 66.58                                      & 67.18                                       & 71.84                                      & 47.53                                   & 5.52                                     & 28.49                                    & 53.43                                   & 65.85                                    \\
			IAL  ~\cite{li2023intensity}            & 87.95                                        & 67.21                                      & 70.10                                       & 76.06                                      & 62.22                                   & 0.00                                     & 36.44                                    & 55.71                                   & 69.24                                    \\
			M3DFEL  ~\cite{wang2023rethinking}      & 89.59                                        & 68.38                                      & 67.88                                       & 74.24                                      & 59.69                                   & 0.00                                     & 31.64                                    & 56.10                                   & 69.25                                    \\
			MAE-DFER ~\cite{sun2023mae}             & 92.92                                        & 77.46                                      & 74.56                                       & 76.94                                      & 60.99                                   & \underline{18.62}                        & \underline{42.35}                        & 63.41                                   & 74.43                                    \\
			SVFAP  ~\cite{sun2023svfap}             & 93.13                                        & 76.98                                      & 72.31                                       & 77.54                                      & \underline{65.42}                       & 15.17                                    & 39.25                                    & 62.83                                   & 74.27                                    \\
			S2D  ~\cite{chen2023static}             & \underline{93.62}                            & \textbf{80.25}                             & \textbf{77.14}                              & \textbf{81.09}                             & 64.53                                   & 1.38                                     & 34.71                                    & 61.82                                   & \underline{76.03}                        \\
			\midrule
			\rowcolor{cyan!10}
			VideoMAE (Baseline)  ~\cite{sun2023mae} & 93.09                                        & 78.78                                      & 71.75                                       & 78.74                                      & 63.44                                   & 17.93                                    & 41.46                                    & \underline{63.60}                       & 74.60                                    \\
			\rowcolor{cyan!10}
			\textbf{S4D (Ours)}                     & \textbf{94.76} (\textcolor{red}{+1.67})      & \underline{79.04} (\textcolor{red}{+0.26}) & \underline{74.60}  (\textcolor{red}{+2.85}) & \underline{79.80} (\textcolor{red}{+1.06}) & \textbf{67.26} (\textcolor{red}{+3.82}) & \textbf{27.59}  (\textcolor{red}{+9.66}) & \textbf{44.57}  (\textcolor{red}{+3.11}) & \textbf{66.80} (\textcolor{red}{+3.20}) & \textbf{76.68}  (\textcolor{red}{+2.08}) \\
			\bottomrule
		\end{tabular}
	}

	\label{tab:specific}
\end{table*}

\subsection{Comparison with the State of the Art}
To evaluate the performance of our S4D method, we compared it with state-of-the-art methods on three publicly available in-the-wild DFER datasets:  DFEW \cite{DFEW}, FERV39K \cite{ferv39k2022}, and MAFW \cite{MAFW}. The results are summarized in Table \ref{tab:sota}.

As shown in Table \ref{tab:sota}, S4D achieves the best performance across all three datasets, significantly outperforming previous state-of-the-art methods, including  A$^3$lign-DFER~\cite{align-dfer}, MAE-DFER~\cite{sun2023mae}, and S2D~\cite{chen2023static}. Specifically, S4D outperforms S2D by 0.65\% WAR on DFEW, 1.09\% WAR on FERV39k, and 1.07\% WAR on MAFW. Additionally, S4D shows substantial improvements in UAR, outperforming S2D by 4.98\% on DFEW, 2.13\% on FERV39k, and 1.92\% on MAFW.
These improvements highlight that \textit{\textbf{our method effectively learns robust representations and mitigates the impact of class imbalance.}}
Compared to the baseline method VideoMAE~\cite{sun2023mae}, S4D achieves significant improvements of +3.20\%/+2.08\% UAR/WAR on DFEW, +0.08\%/+1.26\% UAR/WAR on FERV39k, and +2.87\%/+4.93\% UAR/WAR on MAFW. These results demonstrate S4D’s effectiveness in learning representations through dual-modal pre-training and joint fine-tuning on both static and dynamic FER data, underscoring its superiority in DFER tasks.

Additionally, Table \ref{tab:specific} presents a detailed analysis of S4D's category-specific performance on the DFEW dataset, clearly illustrating its substantial gains across both majority and minority emotion classes.  Specifically, S4D achieves the highest accuracy in \textit{happy} (94.76\%), \textit{surprise} (67.26\%), \textit{disgust} (27.59\%), and \textit{fear} (44.57\%), with a notable 9.66\% improvement in \textit{disgust} over the baseline method VideoMAE \cite{sun2023mae}. It also ranks second in \textit{sad} (79.04\%), \textit{neutral} (74.60\%), and \textit{angry} (79.80\%).  Importantly, S4D demonstrates particularly strong improvements in traditionally underrepresented and challenging categories such as \textit{surprise}, \textit{fear}, and \textit{disgust}, which are often affected by data imbalance.   These improvements are largely attributed to S4D’s dual-modal pre-training and joint fine-tuning framework, which effectively exploits the complementary information from SFER data through unified dual-modal learning, resulting in \textbf{more robust emotion recognition across all categories}.
Although the improvement in \textit{disgust} accuracy is noteworthy, the relatively low absolute accuracy (27.59\%) underscores the inherent difficulty in recognizing this subtle emotion and suggests potential data imbalance within the DFEW dataset. Further investigation into techniques for mitigating data imbalance may yield additional improvements in this category.

\subsection{Ablation Studies}
To evaluate the effectiveness of the key components in the S4D framework, we conduct ablation studies on the FERV39K \cite{ferv39k2022}, DFEW \cite{DFEW}, and AffectNet-7 \cite{Mollahosseini2017AffectNetAD} datasets. For simplicity and to reduce computational costs, we report the results for DFEW based on one fold of the 5-fold cross-validation.

\subsubsection{Impact of Dual-Modal Pre-training}
Table \ref{tab:ablation_pretrain} analyzes the impact of various pre-training settings on DFER performance. All pre-training methods result in significant improvements in both UAR and WAR compared to the no pre-training setting, emphasizing the importance of pre-training in DFER. Specifically, pre-training on single-modal video data (lines 3 and 4) outperforms image-based pre-training, as it captures crucial temporal information. However, image-based pre-training also proves beneficial by enabling the model to learn fine-grained facial features that complement the temporal representations derived from video data. Dual-modal pre-training, which combines image (AffectNet) and video (VoxCeleb2) datasets (line 5), achieves the best performance, with a UAR/WAR of 43.41\%/53.65\% on FERV39K and 70.58\%/79.37\% UAR/WAR on DFEW. This confirms that \textit{\textbf{dual-modal pre-training effectively captures both temporal information and fine-grained facial details}}, leading to robust representation capabilities. Last experiment in line 5, which utilized homogeneous image and video data from VoxCeleb2 for dual-modal pre-training, exhibited slightly lower performance than experiment in line 5. This suggests that \textit{\textbf{leveraging diverse datasets and modalities for pre-training can yield richer and more beneficial feature representations}}. Overall, dual-modal pre-training significantly enhances DFER performance, surpassing both single-modal and no pre-training approaches.

\begin{table}[!t]
	\centering
	\caption{Comparison of the proposed dual-modal pre-training with other pre-training settings. Dual-modal pre-training on AffectNet and VoxCeleb2 significantly improves DFER performance and outperforms single-modal pre-training settings with less pre-training cost. AFC: AffectNet; VC2: VoxCeleb2; $^*$: Equal size with AffectNet.}

	\fontsize{9}{12}\selectfont
	\setlength{\tabcolsep}{1mm} %
	\resizebox{\linewidth}{!}{
		\begin{tabular}{lcccccc}
			\toprule
			\multirow{2}{*}{\tabincell{c}{Pre-Training Setting                                                                                                                                 \\ (images + videos)}}  & \multirow{2}{*}{\tabincell{c}{Pre-Training Cost \\ (TFLOPs)}} & \multicolumn{2}{c}{FERV39K} & \multicolumn{2}{c}{DFEW} \\
			\cmidrule(lr){3-4} \cmidrule(lr){5-6}
			                                       &                                   & UAR                     & WAR                     & UAR                     & WAR                     \\
			\midrule
			Without pre-training                   & 0                                 & 27.67                   & 39.56                   & 40.96                   & 52.65                   \\
			Image only (AFC)                       & $8.8\times10^4$                   & 36.10                   & 47.59                   & 53.80                   & 66.82                   \\
			{Video only (VC2$^*$)} & {$4.6\times10^5$} & {41.11} & {51.78} & {65.92} & {76.41} \\
			Video only (VC2)                       & $1.8\times10^6$                   & 42.92                   & 52.58                   & 67.71                   & 76.33                   \\
			\rowcolor{cyan!10}
			Dual-modal (AFC + VC2)                 & $\bm{1.4\times10^6}$              & \textbf{43.41}          & \textbf{53.65}          & \textbf{70.58}          & \textbf{79.37}          \\
			Dual-modal (VC2$^*$ + VC2)             & $1.4\times10^6$                   & 42.36                   & 52.94                   & 68.35                   & 78.72                   \\
			\bottomrule
		\end{tabular}
	}

	\label{tab:ablation_pretrain}
\end{table}
\subsubsection{Pre-Training Cost Analysis}
As shown in Table \ref{tab:ablation_pretrain}, the proposed dual-modal pre-training (line 5) demonstrates superior efficiency and effectiveness, significantly reducing computational costs compared to video-only pre-training (line 3). Specifically, the integration of image data reduces the pre-training cost by approximately 22\% (from 1.8 to 1.4 $\times$ 10$^6$ TFLOPs), achieved through aggressive masking ratios of 90\% for images and 95\% for videos, minimizing redundant computations. This approach not only optimizes efficiency but also enhances performance, yielding notable improvements on the FERV39K (+0.49\% UAR, +1.07\% WAR) and DFEW (+2.87\% UAR, +3.04\% WAR) benchmarks. In contrast, image-only pre-training (line 2) incurs a minimal cost (8.8 $\times$ 10$^4$ TFLOPs) but shows limited effectiveness for DFER tasks. Similarly, the no pre-training setting (line 1) incurs no pre-training cost but severely degrades performance.  \textit{\textbf{These findings underscore the critical importance of capturing temporal dynamics for DFER}, which are absent in image-only pre-training or no pre-training strategies.}
\begin{table}[!t]
	\centering
	\caption{Performance and parameter efficiency of S4D compared to other multi-task learning (MTL) methods. ViT-MTL refers to the direct application of ViT in MTL tasks with multiple heads. ViT-MoE \cite{NEURIPS2021_48237d9f} replaces the FFN in the ViT layer with an MoE and employs FFNs as experts.}

	\setlength{\tabcolsep}{1mm} %
	\resizebox{\linewidth}{!}{
		\begin{tabular}{lccccc}
			\toprule
			\multirow{2}{*}{Model Setting} & Params (M)        & \multicolumn{2}{c}{FERV39K} & \multicolumn{2}{c}{DFEW}                                   \\
			\cmidrule(lr){3-4} \cmidrule(lr){5-6}
			                               & Test / Train      & UAR                         & WAR                      & UAR            & WAR            \\
			\midrule
			ViT                            & 86 / 86           & 42.68                       & 53.08                    & 68.69          & 77.01          \\
			ViT-MTL (w/o MoAE)             & 86 / 86           & 42.63                       & 52.96                    & 69.62          & 78.08          \\
			\rowcolor{cyan!10}     %
			S4D (Ours)                     & \textbf{90 / 101} & \textbf{43.41}              & \textbf{53.65}           & \textbf{70.58} & \textbf{79.37} \\
			\midrule
			ViT-MoE                        & 115 / 285         & 41.79                       & 52.81                    & 68.88          & 78.12          \\
			\bottomrule
		\end{tabular}
	}
	\label{tab:ablation_on_moae}
\end{table}

\subsubsection{Evaluation of the MoAE Module}

To evaluate MoAE's effectiveness in mitigating negative transfer between SFER and DFER tasks, we conducted comprehensive comparisons with other multi-task learning (MTL) methods. As shown in Table \ref{tab:ablation_on_moae}, the vanilla ViT model achieves a performance of 42.68\%/53.08\% UAR/WAR  on FERV39K and 68.69\%/77.01\% UAR/WAR on DFEW dataset. When SFER data is incorporated through simple multi-task learning (simply adding multiple task-specific heads to ViT, line 3), we observe minimal improvement or even slight performance degradation compared to the vanilla ViT. This phenomenon suggests \textit{\textbf{that straightforward multi-task adaptation may fail to capture task relationships and lead to negative transfer.}} In contrast, our proposed S4D (with MoAE) demonstrates significant improvements across all metrics while maintaining reasonable parameter efficiency. Specifically, S4D achieves the best performance with 43.41\%/53.65\% UAR/WAR on FERV39K and 70.58\%/79.37\% UAR/WAR on DFEW, validating its effectiveness in learning both task-specific knowledge and generic features.
To further investigate the architectural benefits, we compared S4D with ViT-MoE \cite{NEURIPS2021_48237d9f}, a classic MoE architecture that replaces the standard FFN with FFN experts. Despite utilizing substantially more parameters (115M/285M vs. 90M/101M for testing/training), ViT-MoE exhibits poorer performance. These results emphasize that \textit{\textbf{S4D's superior performance stems from its architectural innovations rather than mere parameter scaling}, highlighting its parameter efficiency and effectiveness in learning complementary knowledge between SFER and DFER tasks while mitigating negative transfer}.

\begin{table}[!t]
	\centering
	\caption{Ablation studies on the joint fine-tuning strategy. {Joint fine-tuning improves DFER significantly without notably compromising SFER. *: Results obtained at epoch 43, which yields optimal SFER performance. Other configurations are selected based on best DFER metrics.}}

	\setlength{\tabcolsep}{1mm}
	\resizebox{\linewidth}{!}{
		\begin{tabular}{lcccc}
			\toprule
			\multirow{2}{*}{ \tabincell{c}{Fine-Tuning Setting                                                          \\(data mixture) }}             & \multicolumn{2}{c}{FERV39K} & \multicolumn{2}{c}{AffectNet-7}                                   \\
			\cmidrule(lr) {2-3} \cmidrule(lr) {4-5}

			                                        & UAR            & WAR            & UAR            & WAR            \\
			\midrule
			SFER (Static)                           & -              & -              & 65.58          & 65.64          \\

			DFER  (Dynamic)                         & 42.45          & 52.82          & -              & -              \\
			\rowcolor{cyan!10}     %
			Joint (50\% Static + 100\% Dynamic)     & \textbf{43.41} & \textbf{53.65} & 65.45          & 65.50          \\
			Joint (50\% Static + 100\% Dynamic)$^*$ & 40.52          & 50.90          & \textbf{65.88} & \textbf{65.94} \\

			\bottomrule
		\end{tabular}
	}
	\label{tab:ablation_on_joint_learning}
\end{table}

\subsubsection{The Effectiveness of Joint Fine-Tuning}

We conducted an ablation study to assess the effectiveness of joint fine-tuning on both SFER and DFER tasks, as presented in Table \ref{tab:ablation_on_joint_learning}. Our findings reveal that joint fine-tuning significantly outperforms individual fine-tuning, particularly for DFER. On the FERV39K dataset, joint fine-tuning improves UAR/WAR by 0.96\%/0.83\% over DFER-only fine-tuning. For SFER on the AffectNet-7 dataset, it improves UAR/WAR by 0.30\%/0.30\%. \textit{These results underscore \textbf{the complementary nature of static and dynamic expression knowledge and demonstrate the effectiveness of joint fine-tuning} in enhancing expression representation learning}. They also highlight S4D's ability to handle both SFER and DFER tasks simultaneously, effectively leveraging the strengths of each modality. However, achieving optimal results required different training epochs for DFER and SFER, which indicates challenges in simultaneous optimization.

\begin{table}[!t]
	\centering
	\caption{Contribution of proposed components. D.P.: Dual-Modal Pre-training, J.F.: Joint Fine-tuning.}

	\resizebox{\linewidth}{!}{
		\begin{tabular}{ccccccc}
			\toprule
			\multirow{2}{*}{D.P.}        & \multirow{2}{*}{MoAE} & \multirow{2}{*}{J.F.}        & \multicolumn{2}{c}{FERV39K} & \multicolumn{2}{c}{DFEW}                                                     \\
			\cmidrule(lr) {4-5} \cmidrule(lr) {6-7}
			                             &                       &                              & UAR                         & WAR                      & UAR                     & WAR                     \\
			\midrule
			                             &                       &                              & 42.58                       & 52.34                    & 64.66                   & 76.16                   \\
			\checkmark                   &                       &                              & 42.68                       & 53.08                    & 68.69                   & 77.01                   \\
			\checkmark                   & \checkmark            &                              & 42.45                       & 52.82                    & 66.52                   & 76.93                   \\
			{\checkmark} & {}    & {\checkmark} & {42.63}     & {52.96}  & {69.62} & {78.08} \\
			\rowcolor{cyan!10}     %
			\checkmark                   & \checkmark            & \checkmark                   & \textbf{43.41}              & \textbf{53.65}           & \textbf{70.58}          & \textbf{79.37}          \\
			\bottomrule
		\end{tabular}
	}
	\label{tab:combined_table}
\end{table}
\subsubsection{Contribution of Proposed Components}
We conducted ablation studies to evaluate the impact of each component in our S4D framework: dual-modal pre-training (D.P.), MoAE, and joint fine-tuning (J.F.). Results from Table \ref{tab:combined_table} indicate that D.P. improved UAR/WAR by 0.10\%/0.74\% on FERV39K and 4.03\%/0.85\% on DFEW over the baseline (our fine-tuned version of VideoMAE on the pre-trained weights from \cite{sun2023mae}), underscoring its role in enhancing spatiotemporal representations. However, MoAE alone led to a slight performance decrease on all metrics, suggesting it may require a joint fine-tuning strategy to unlock its task-specific capabilities fully. The optimal configuration, integrating all three components {(dual-modal pre-training, joint fine-tuning, and the MoAE module), achieved  UAR/WAR of 43.41\%/53.65\% on FERV39K and 70.58\%/79.37\% on DFEW. These results highlight the synergistic effect of the full S4D framework, where the integration of data diversity, task coupling, and architectural enhancement collectively drives substantial improvements in DFER performance.}

\subsubsection{Ablation Study on the Position of MoAE}
To identify the optimal placement of the MoAE module within the ViT architecture, we conducted ablation studies at different positions.
Table \ref{tab:combined_table2} shows that placing the MoAE in the last six layers (Later) yields the highest UAR/WAR performance at 43.41\%/53.65\% on FERV39K and 70.58\%/79.37\% on DFEW, respectively. In contrast, placing the MoAE in the first six layers (Early) results in a lower UAR/WAR of 43.15\%/53.46\% on FERV39K and 67.93\%/78.42\% on DFEW, suggesting that early placement may hinder the model’s ability to learn generic features. Middle and alternate layer placements produce intermediate results. These findings indicate that {the MoAE module is most effective in the later layers, where it can enhance task-specific learning by leveraging the learned generic representations. Conversely, placing them too early could disrupt the formation of these foundational generic features.}

\begin{table}[!t]
	\centering
	\caption{Ablation studies on the position of the MoAE module within the ViT architecture.}

	\resizebox{0.8\linewidth}{!}{
		\begin{tabular}{lcccc}
			\toprule
			\multirow{2}{*}{ Position} & \multicolumn{2}{c}{FERV39K} & \multicolumn{2}{c}{DFEW}                                   \\
			\cmidrule(lr) {2-3} \cmidrule(lr) {4-5}

			                           & UAR                         & WAR                      & UAR            & WAR            \\
			\midrule
			Early                      & 43.15                       & 53.46                    & 67.93          & 78.42          \\

			Middle                     & 43.32                       & 53.57                    & 69.96          & 79.15          \\
			\rowcolor{cyan!10}     %
			Later                      & \textbf{43.41}              & \textbf{53.65}           & \textbf{70.58} & \textbf{79.37} \\
			Alternate                  & 43.27                       & 53.52                    & 70.48          & 79.20          \\

			\bottomrule
		\end{tabular}
	}
	\label{tab:combined_table2}
\end{table}

\begin{table}[!t]
	\centering
	\caption{Ablation Studies on the Masking Ratios. AFC: AffectNet; VC2: VoxCeleb2}
	\resizebox{0.95\linewidth}{!}{
		{
				\begin{tabular}{cc|cccc}
					\toprule
					\multicolumn{2}{c|}{Masking Ratio} & \multicolumn{2}{c}{FERV39K} & \multicolumn{2}{c}{DFEW}                                                    \\
					\cmidrule(lr){1-6}
					AFC                                & VC2                         & UAR                      & WAR            & UAR            & WAR            \\
					\midrule

					75\%                               & 75\%                        & 42.03                    & 52.50          & 65.43          & 76.97          \\
					75\%                               & 90\%                        & 43.43                    & 52.67          & 65.95          & 77.27          \\
					75\%                               & 95\%                        & 42.54                    & 52.47          & 66.65          & 77.35          \\
					90\%                               & 75\%                        & 42.28                    & 52.64          & 68.51          & 78.25          \\

					90\%                               & 90\%                        & 42.82                    & 52.96          & 68.10          & 78.08          \\
					\rowcolor{cyan!10}     %
					90\%                               & 95\%                        & \textbf{43.41}           & \textbf{53.65} & \textbf{70.58} & \textbf{79.37} \\

					\bottomrule
				\end{tabular}
			}
	}
	\label{tab:mask_ratio}
\end{table}

\subsubsection{{Ablation Study on the Masking Ratios}}
{To assess the impact of masking ratios, we perform an ablation study during pre-training on AffectNet (images) and VoxCeleb2 (videos). As presented in Table~\ref{tab:mask_ratio}, the combination of 90\% image and 95\% video masking ratios consistently yields the best performance, achieving 43.41\%/53.65\% UAR/WAR on FERV39K and 70.58\%/79.37\% on DFEW. Notably, this high masking ratio significantly reduces the computational cost during pre-training. Specifically, it reduces FLOPs by approximately 42\% for image inputs compared to the MAE \cite{he2022masked} with a 75\% masking ratio, and by 22\% for video inputs compared to VideoMAE  \cite{tong2022videomae} with a 90\% masking ratio. These findings indicate that an aggressive masking strategy is not only effective in performance but also computationally efficient for fine-grained tasks such like DFER. Based on this analysis, we adopt a 90\% masking ratio for images and a 95\% masking ratio for videos in our final pre-training models.}

\begin{figure}[!t]
	\centering
	\begin{subfigure}[t]{0.48\linewidth}
		\centering
		\includegraphics[width=\textwidth]{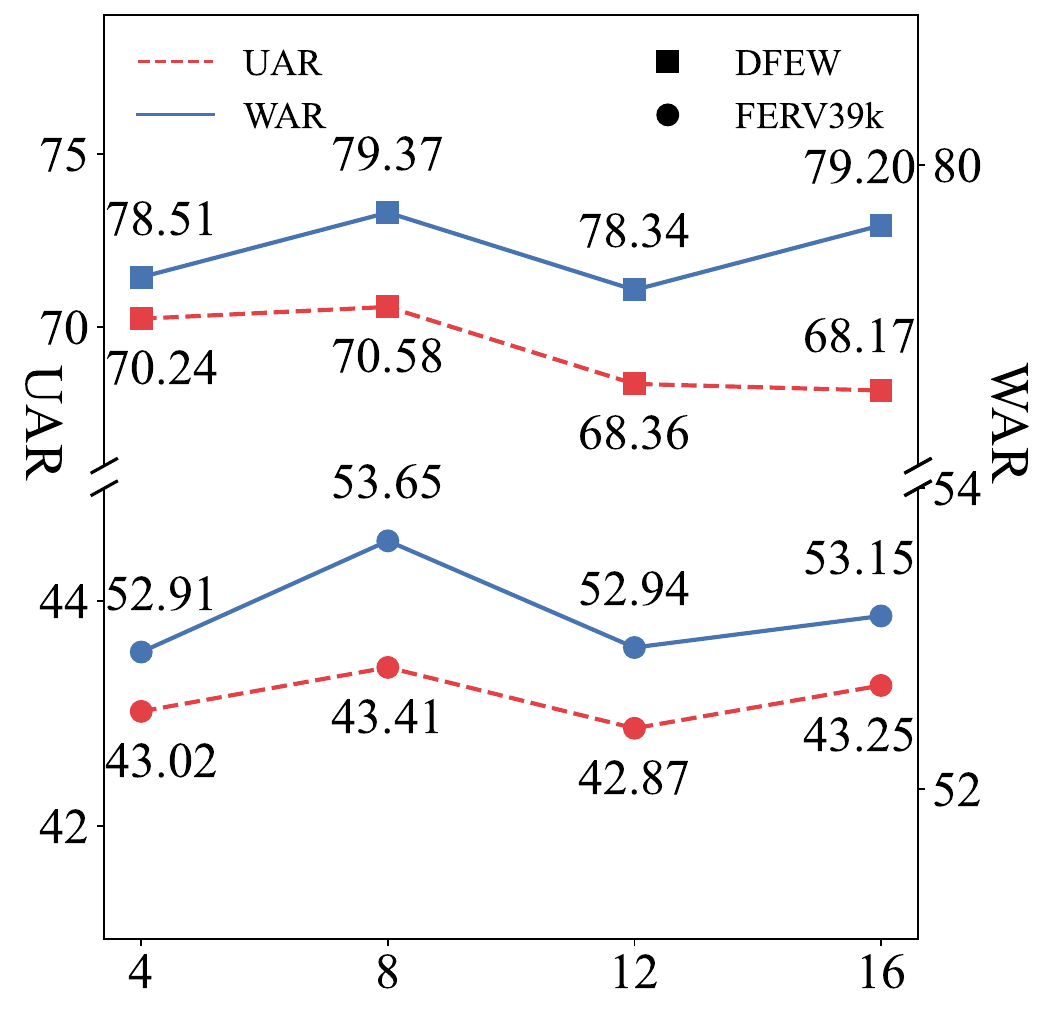}
		\caption{Total number of experts $n$.} \label{fig:ablation_num_moe_experts}
	\end{subfigure}\hfill
	\begin{subfigure}[t]{0.48\linewidth}
		\centering
		\includegraphics[width=\textwidth]{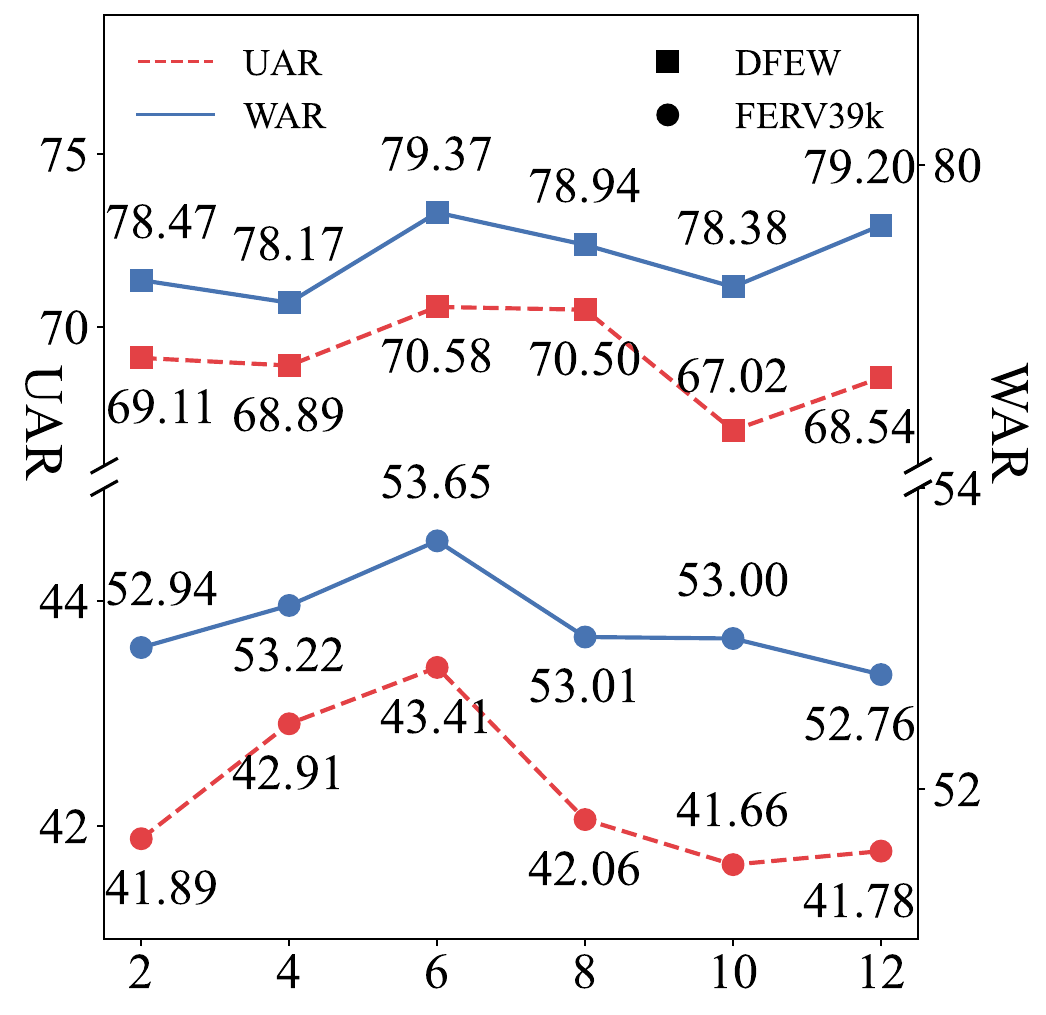}
		\caption{The number of MoAE layers.}
		\label{fig:ablation_num_moe_layers}
	\end{subfigure}
	\begin{subfigure}[t]{0.48\linewidth}
		\centering
		\includegraphics[width=\textwidth]{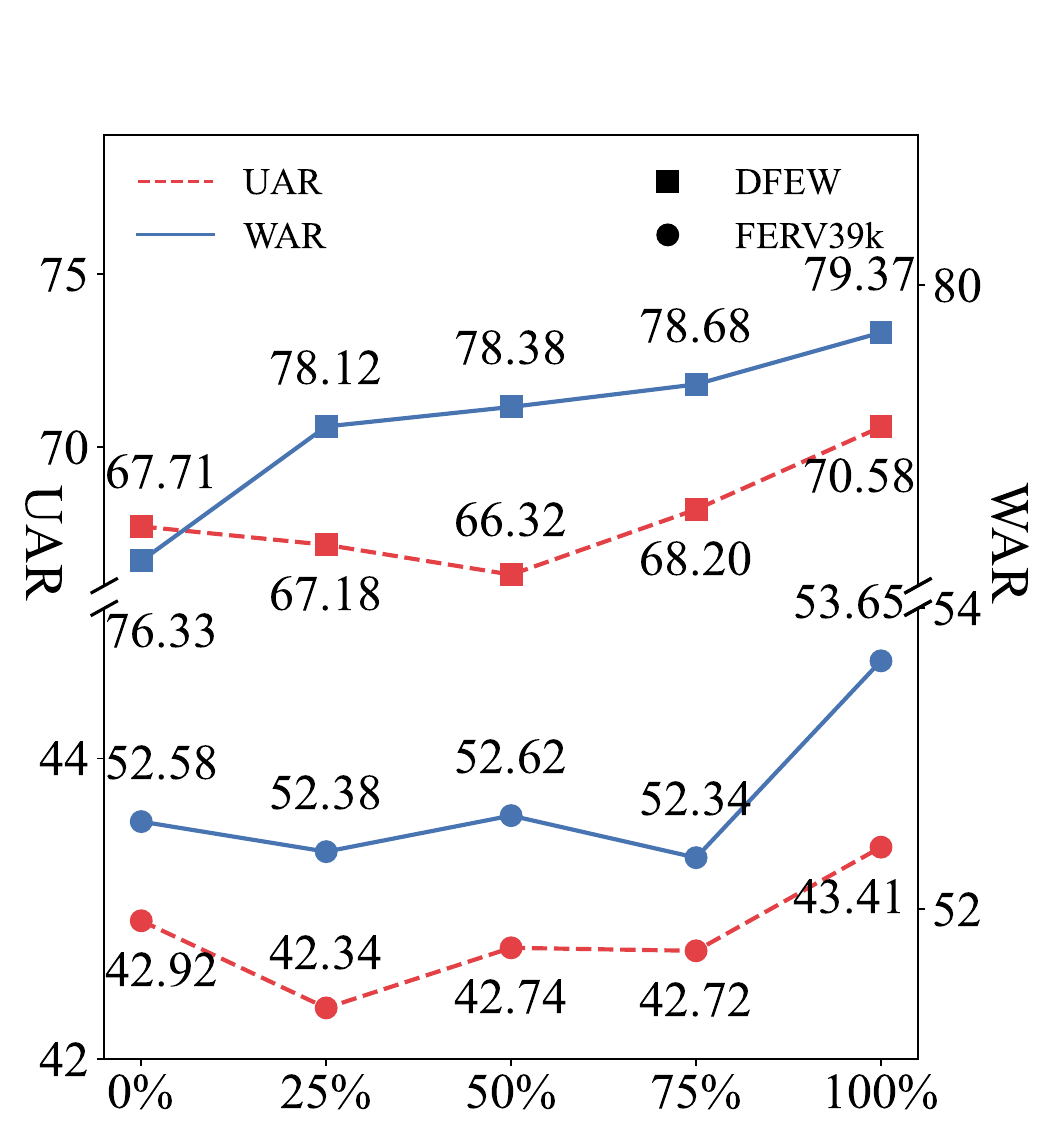}
		\caption{Proportion of SFER dataset used during pre-training.}
		\label{fig:ablation_pretrain_data}
	\end{subfigure}\hfill
	\begin{subfigure}[t]{0.48\linewidth}
		\centering
		\includegraphics[width=\textwidth]{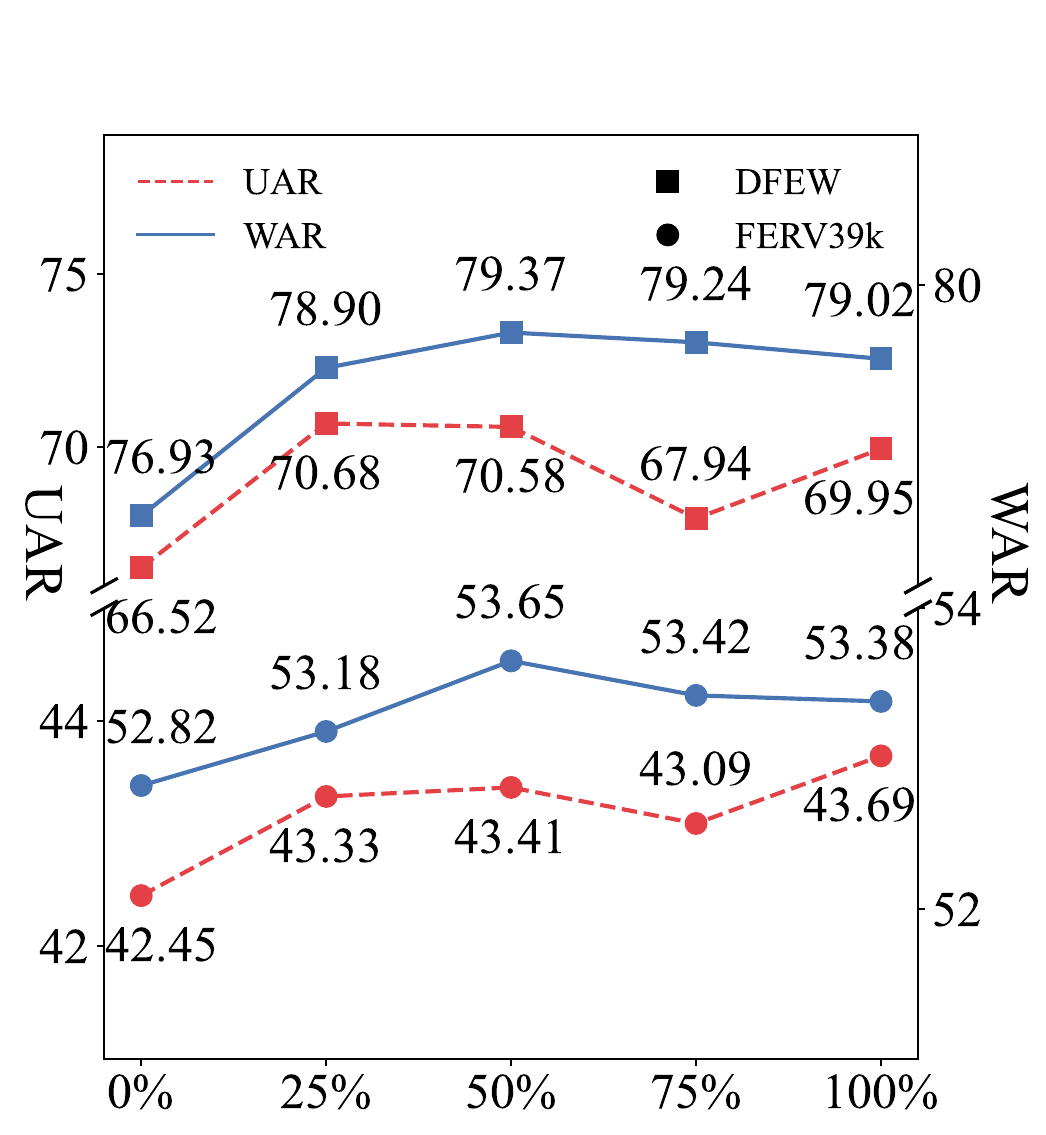}
		\caption{Proportion of SFER dataset used during fine-tuning.}
		\label{fig:ablation_finetune_data}
	\end{subfigure}

	\caption{Analyses of the total number of experts, the number of MoAE layers, and the proportion of SFER data used during dual-modal pre-training and joint fine-tuning.}
	\label{fig:ablation_on_hp2}
\end{figure}

\subsubsection{Ablation Studies on Hyperparameters}

We conducted comprehensive ablation studies to investigate the impact of three critical hyperparameters: the number of experts ($n$) in the MoAE module, the number of MoAE layers, and the proportion of the SFER dataset utilized during dual-modal pre-training and joint fine-tuning.

Our experiments first focused on the structural parameters of the MoAE module. The results indicate that the model achieves optimal performance in both the FERV39K and DFEW datasets when the MoAE module is configured with 8 experts, as illustrated in Fig. \ref{fig:ablation_num_moe_experts}. Any deviation from this optimal number, whether an increase or decrease, results in marginal performance degradation. {This suggests that 8 experts strike an effective balance between sufficient model capacity to learn diverse features and mitigating the risks of overfitting or excessive computational overhead that might arise from a larger number of experts. }Similarly, our investigation into the impact of MoAE layer count reveals that incorporating six MoAE layers yields the best performance, as shown in Fig. \ref{fig:ablation_num_moe_layers}. Deviating from this number leads to a noticeable decline in performance.  These findings suggest that replacing the final six ViT layers with MoAE layers establishes an optimal balance between task-specific adaptation and preservation of general knowledge.

Beyond structural parameters, we also explored the impact of external training data.
As depicted in Fig. \ref{fig:ablation_pretrain_data}, increasing the proportion of the SFER dataset during the pre-training phase leads to consistent performance improvements on the DFEW dataset (up to 2.87\%/3.04\% UAR/WAR gain), whereas the impact on FERV39k is less pronounced (approximately 0.49\%/1.07\% UAR/WAR improvement). These differences may reflect variations in dataset characteristics, such as scale, label noise, or inherent task difficulty. Both datasets achieve peak performance when the entire SFER dataset is employed during pre-training, indicating that pre-training on SFER data is generally advantageous, especially for DFEW. Regarding the SFER dataset utilization during fine-tuning, our experiments reveal that {increasing the proportion of SFER data from scratch enhances performance. Notably, using 50\% of the SFER data yields optimal results, as illustrated in Fig. \ref{fig:ablation_finetune_data}. However, further increasing the proportion beyond this point} offers no additional benefits and could bias the optimization trajectory of the DFER task, {potentially leading to  a degradation in DFER performance.}

Overall, these empirical findings highlight the importance of careful hyperparameter tuning in maximizing the effectiveness of the proposed unified learning framework. Specifically, using 8 MoAE experts, six MoAE layers, and the entire SFER dataset during pre-training, along with partial SFER data during fine-tuning, establishes a robust foundation for practical deployment while maintaining computational efficiency.

\begin{table}[!t]
	\centering
	\caption{{Ablation studies on the number of activated experts per MoAE layer.}}
	{
		\begin{tabular}{ccccc}
			\toprule
			\multirow{2}{*}{ \tabincell{c}{Activated                                          \\Experts}}             & \multicolumn{2}{c}{FERV39K} & \multicolumn{2}{c}{DFEW}                                   \\
			\cmidrule(lr) {2-3} \cmidrule(lr) {4-5}

			  & UAR               & WAR               & UAR               & WAR               \\
			\midrule

			0 & 42.63             & 52.96             & \underline{69.62} & 78.08             \\
			\rowcolor{cyan!10}
			2 & 43.41             & \textbf{53.65}    & \textbf{70.58}    & \textbf{79.37}    \\
			4 & \textbf{43.38}    & 53.03             & 67.92             & 78.38             \\
			6 & 42.98             & \underline{53.50} & 67.51             & 77.95             \\
			8 & \underline{43.55} & 53.00             & 67.96             & \underline{78.47} \\

			\bottomrule
		\end{tabular}}
	\label{tab:activated_experts}
\end{table}

\begin{table}[!t]
	\centering
	\caption{{Performance comparison of different curriculum strategies for incorporating SFER data during joint fine-tuning. Linear: gradually increasing the proportion of SFER data; Step: introducing SFER data only in the second half of training; Fixed: maintaining a constant 50\% SFER proportion throughout.}}
	{
		\begin{tabular}{ccccc}
			\toprule
			\multirow{2}{*}{ \tabincell{c}{Adjusting                                         \\ Strategy}}             & \multicolumn{2}{c}{FERV39K} & \multicolumn{2}{c}{DFEW}                                   \\
			\cmidrule(lr) {2-3} \cmidrule(lr) {4-5}

			             & UAR            & WAR            & UAR            & WAR            \\
			\midrule

			Linear       & 41.78          & 52.66          & 66.52          & 76.76          \\

			Step         & 41.81          & 52.87          & 66.66          & 77.01          \\
			\rowcolor{cyan!10}
			Fiexd (ours) & \textbf{43.38} & \textbf{53.65} & \textbf{70.58} & \textbf{79.37} \\

			\bottomrule
		\end{tabular}
	}
	\label{tab:adjusting_strategy}
\end{table}

{
	\subsubsection{The Impact of Number Activated Experts}
	Ablation results in Table~\ref{tab:activated_experts} demonstrate that activating 2 experts per MoAE layer yields the best overall performance, achieving the highest WAR on both FERV39K (53.65\%) and DFEW (79.37\%), as well as the highest UAR on DFEW (70.58\%). Compared to the zero-expert setting, which essentially disables expert routing, the inclusion of even two experts significantly improves performance, confirming the functional value of our MoAE module. However, increasing the number of activated experts beyond 2 not only fails to improve performance but also introduces additional computational overhead, suggesting that excessive expert activation introduces redundancy and harms generalization. These findings indicate that a sparse expert activation strategy (i.e., top-2) provides the optimal balance between representational capacity and the risk of overfitting.}
{
	\subsubsection{Comparison of SFER Data Adjusting Strategies}
	Ablation results in Table~\ref{tab:adjusting_strategy} compare three curriculum strategies for incorporating SFER data into training. The proposed Fixed strategy, which maintains a constant 50\% SFER proportion throughout training, outperforms both Linear (gradual increase) and Step (late introduction) baselines, achieving the best UAR/WAR on both FERV39K (43.38\%/53.65\%) and DFEW (70.58\%/79.37\%). Fixed integration allows the model to leverage semantic diversity from the outset, fostering better generalization and mitigating early overfitting to target-specific patterns, whereas delayed (Step) or gradual (Linear) incorporation limits the influence of auxiliary signals influence during the critical representation learning phase. 
}
\setlength{\belowcaptionskip}{-5pt} %

\begin{table}[!t]
	\centering
	\caption{Cross-task evaluation comparisons on SFER (AffectNet-7) and DFER (FERV39K, DFEW).}
	\begin{tabular}{llcc}
		\toprule
		Train       & Test        & UAR   & WAR   \\
		\midrule
		AffectNet-7 & FERV39K     & 32.34 & 35.10 \\
		AffectNet-7 & DFEW        & 43.36 & 47.13 \\
		FERV39K     & AffectNet-7 & 51.46 & 51.38 \\
		DFEW        & AffectNet-7 & 45.20 & 45.04 \\
		\bottomrule
	\end{tabular}
	\label{tab:cross_test}
\end{table}

\setlength{\belowcaptionskip}{-5pt} %
\begin{figure}
	\centering
	\includegraphics[width=\linewidth]{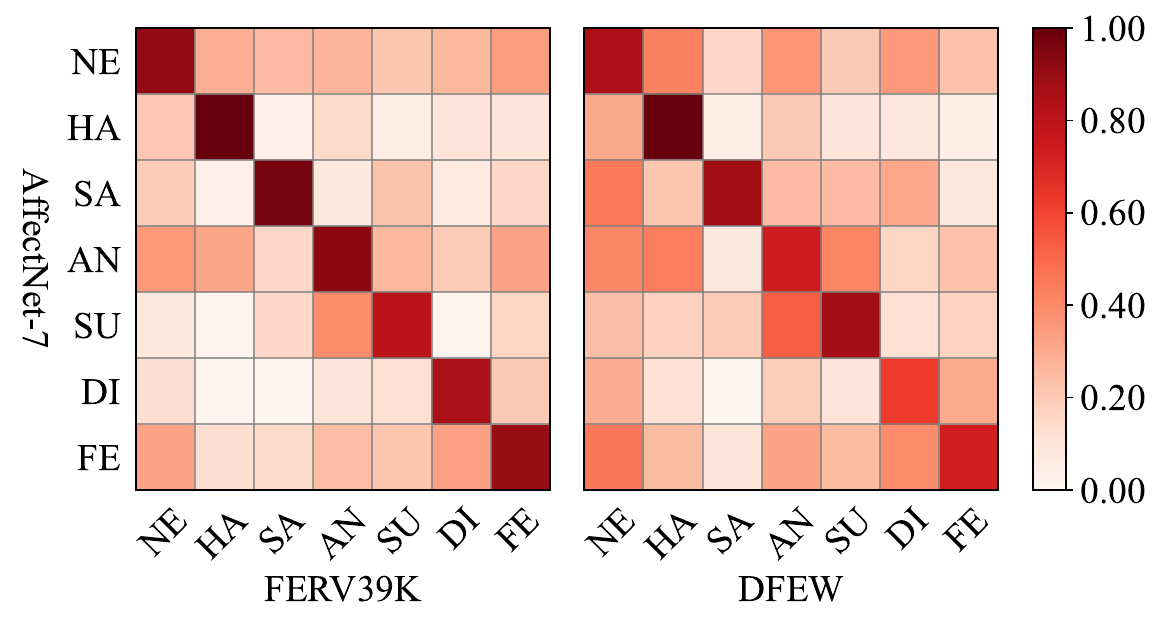}
	\caption{The semantic relevance between SFER and DFER tasks. NE, HA, SA, AN, SU, DI, and FE denote \textit{neutral}, \textit{happy}, \textit{sad}, \textit{anger}, \textit{surprise}, \textit{disgust}, and \textit{fear}, respectively.}
	\label{fig:similarity}
\end{figure}
\setlength{\belowcaptionskip}{-5pt} %

\subsection{Correlation Analysis between SFER and DFER }

\subsubsection{Evaluation on Cross-Task } %
Since S4D can handle both SFER and DFER tasks using a shared backbone and respective classifiers, we treat it as a distinct task model during inference. To investigate the correlations and differences between the SFER and DFER tasks, we conducted cross-task evaluation comparisons on the AffectNet-7, FERV39K, and DFEW datasets. As shown in Table \ref{tab:cross_test}, the SFER model trained on AffectNet-7 achieves 32.34\%/35.10\% and 43.36\%/47.13\% UAR/WAR on the FERV39K and DFEW datasets, respectively.
Although this performance is lower than that of DFER models on their respective tasks, it significantly exceeds random guessing (approximately 14.3\%), indicating that the SFER model captures relevant spatial features from video data. In contrast, DFER models trained on the FERV39K and DFEW datasets achieve 51.46\%/51.38\% and 45.20\%/45.04\% UAR/WAR on AffectNet-7, respectively. These results suggest that DFER models can leverage static expression information from dynamic data, transferring well to the SFER task. We attribute this phenomenon to the inherent presence of static facial expressions within dynamic sequences, which capture peak expressions and serve as snapshots within the temporal progression of the video.
However, direct comparisons of UAR/WAR metrics between SFER and DFER models are not ideal due to the inherent differences between the tasks. Therefore, these results should be viewed as reflecting feature transferability rather than a direct performance comparison.
\setlength{\belowcaptionskip}{-5pt} %

\begin{figure}[!t]
	\centering
	\begin{subfigure}[t]{\linewidth}
		\centering
		\includegraphics[width=\linewidth]{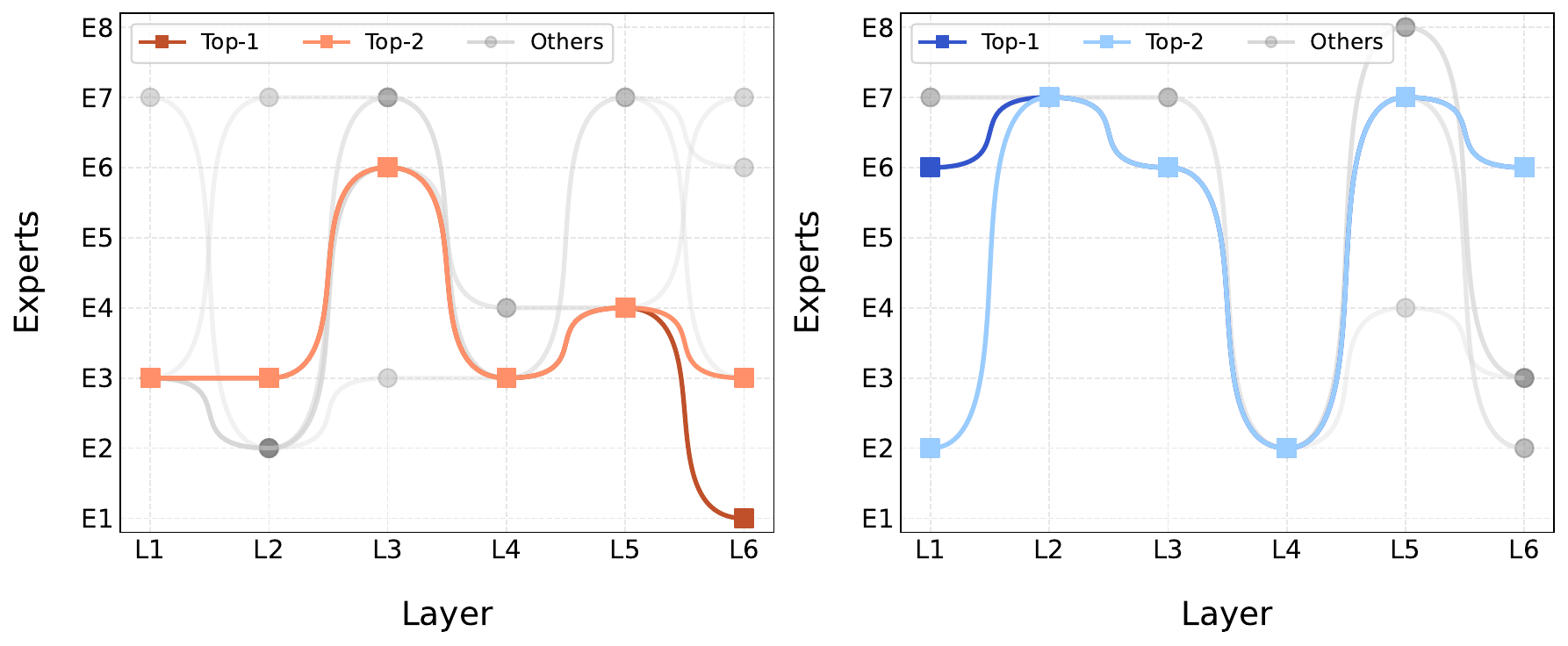}
		\caption{FERV39K vs. AffectNet-7.}
		\label{fig:activation_paths_ferv39k}
	\end{subfigure}
	\vspace{2mm} %

	\begin{subfigure}[t]{\linewidth}
		\centering
		\includegraphics[width=\linewidth]{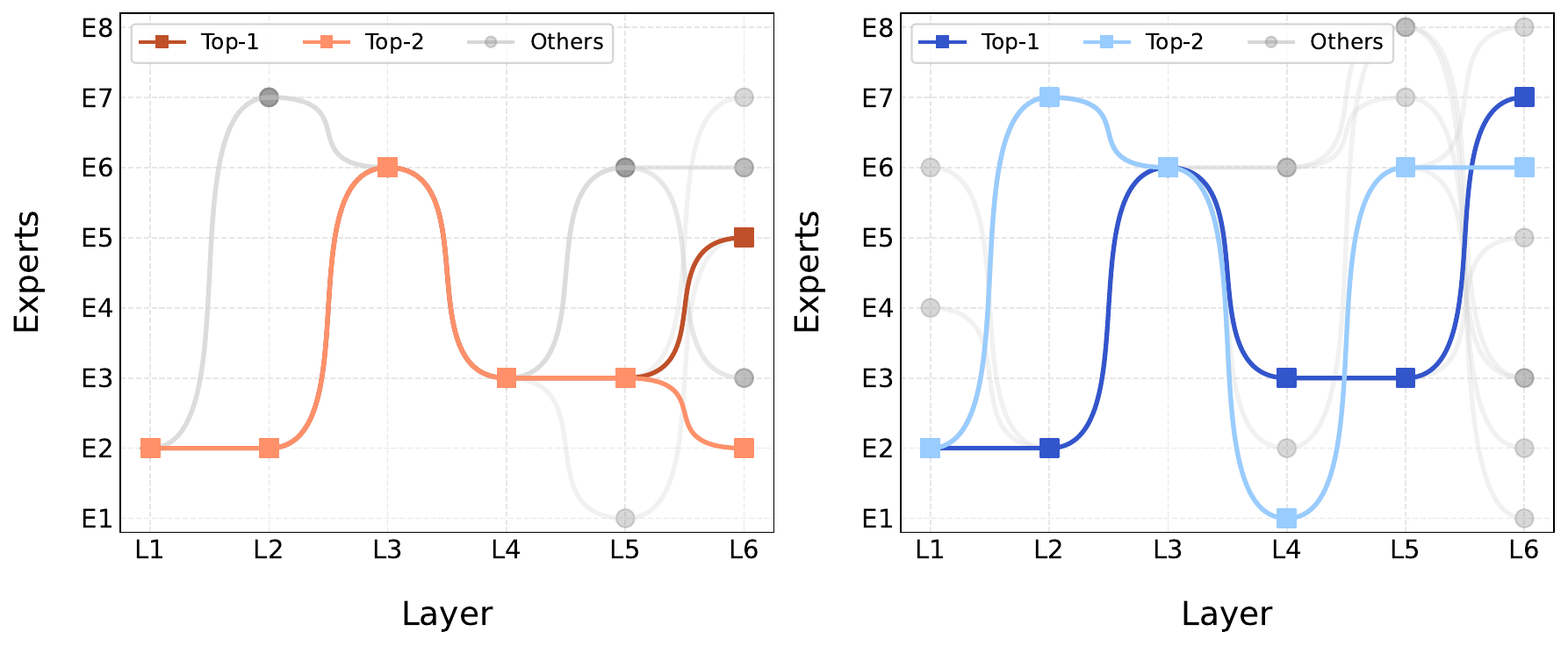}
		\caption{DFEW vs. AffectNet-7.}
		\label{fig:activation_paths_dfew}
	\end{subfigure}

	\caption{
		{Visualization of activation pathways. The figure shows the top-10 most frequently activated expert paths, where the top-2 are highlighted in color and the others in gray. Note that DFER and SFER tasks employ hard sharing through the FFN branch, enabling structural coupling while maintaining task-specific routing.}
	}

	\label{fig:expert_activation_paths}
\end{figure}

\subsubsection{Semantic Relevance between SFER and DFER Tasks}
\begin{figure*}
	\centering
	\includegraphics[width=\linewidth]{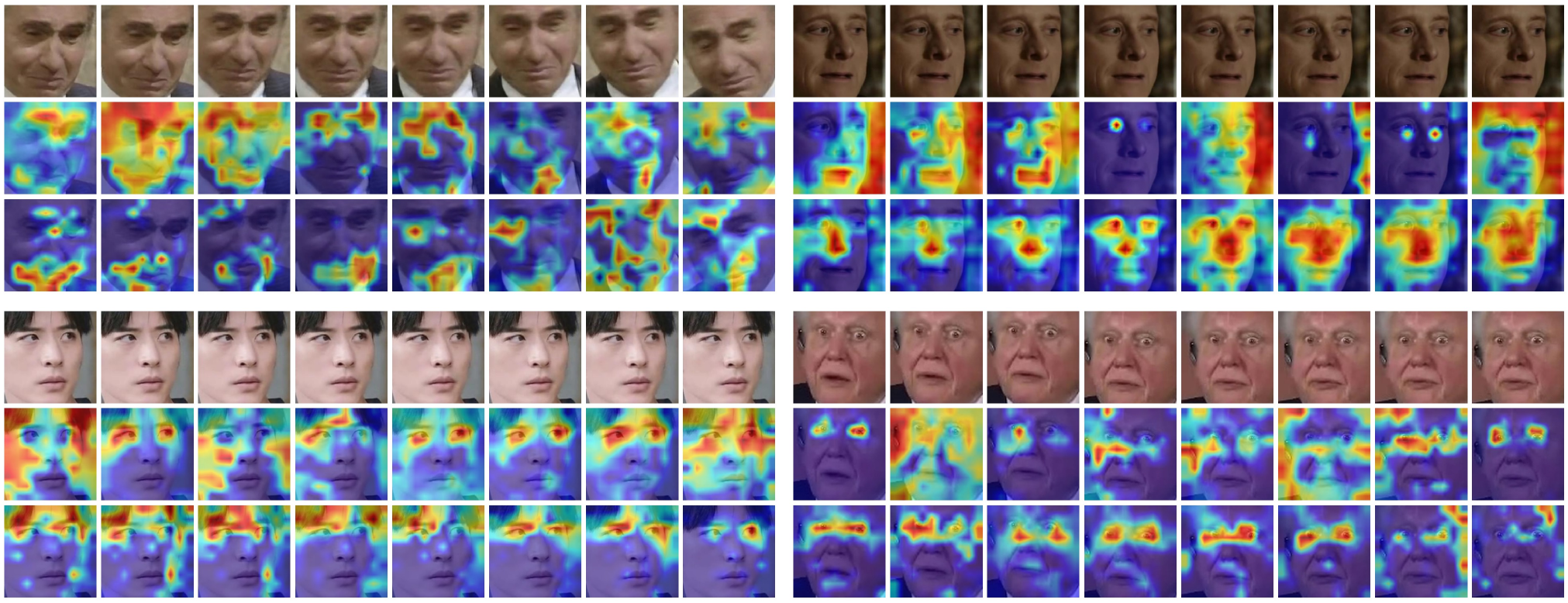}
	\caption{Visualizations of original frames (first row) and attention maps (only positive values) from baseline model (second row) and S4D (third row) across four emotion categories: \textit{sad}, \textit{fear}, \textit{angry}, \textit{surprise} (from left to right, top to bottom). }
	\label{fig:attention_map}
\end{figure*}

To further investigate the semantic relevance between SFER and DFER tasks, we analyzed the cosine similarity between the class representation centers of FERV39K and AffectNet-7, as well as between DFEW and AffectNet-7. As illustrated in Fig. \ref{fig:similarity}, \textit{\textbf{the similarity matrix reveals prominently high values along the diagonal, indicating strong semantic correlations between corresponding expression classes across the two tasks}.}  These findings provide robust empirical evidence supporting our motivation to utilize SFER data to enhance the understanding of dynamic expressions and improve DFER performance.  Moreover, additional regions of elevated similarity were observed, particularly related to the \textit{neutral} expression category, suggesting inherent semantic overlaps among certain emotion categories.

\begin{figure*}[!t]
	\centering
	\begin{subfigure}[b]{0.32\linewidth}
		\centering
		\includegraphics[width=\linewidth]{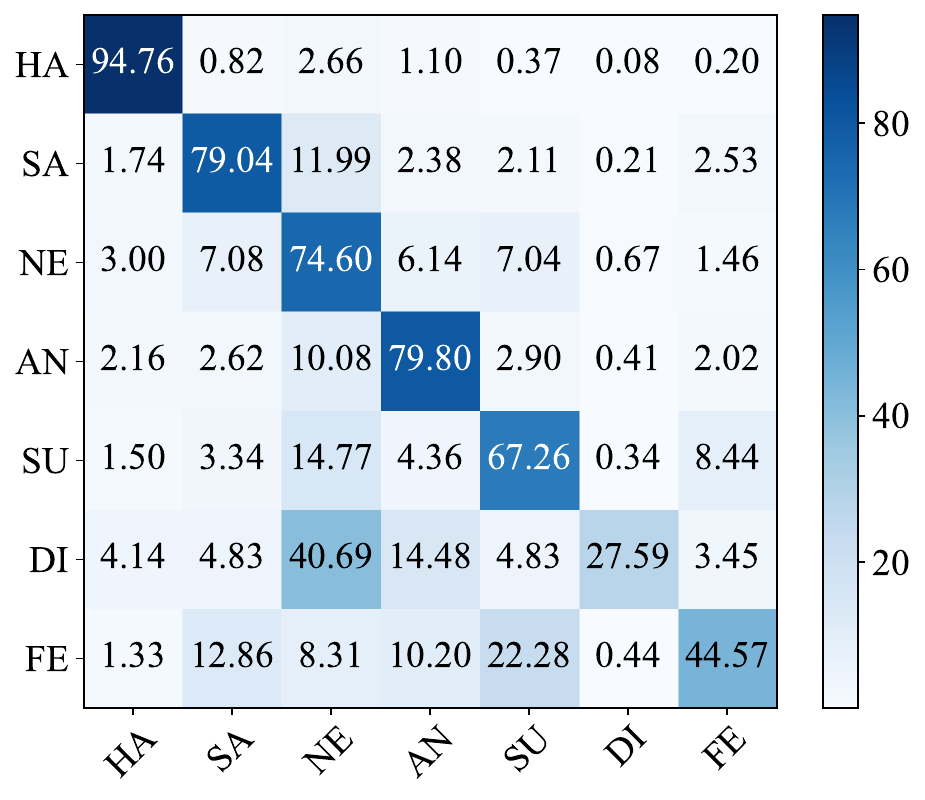}
		\caption{DFEW}
	\end{subfigure}
	\hfill
	\begin{subfigure}[b]{0.32\linewidth}
		\centering
		\includegraphics[width=\linewidth]{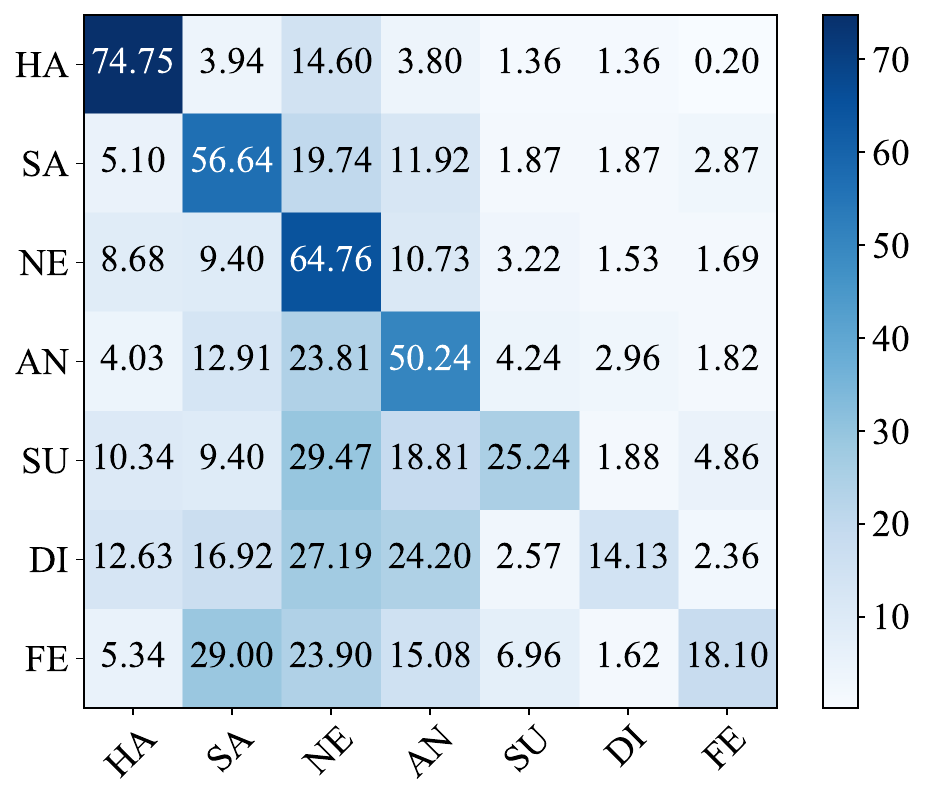}
		\caption{FERV39K}
	\end{subfigure}
	\hfill
	\begin{subfigure}[b]{0.32\linewidth}
		\centering
		\includegraphics[width=\linewidth]{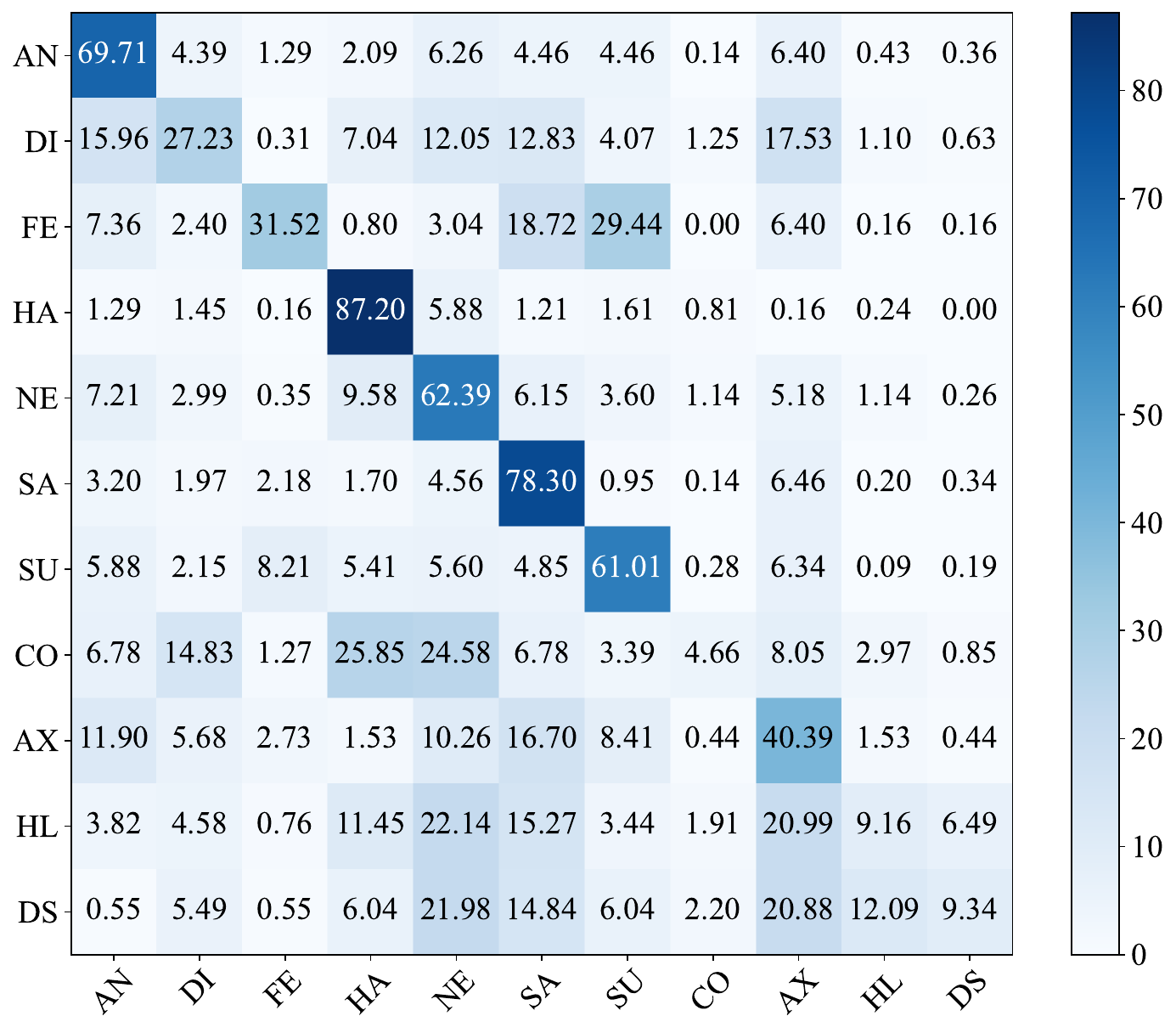}
		\caption{MAFW}
	\end{subfigure}
	\caption{Confusion matrices of S4D on DFEW, FERV39K and MAFW datasets. AN: \textit{angry}. DI: \textit{disgust}. FE: \textit{fear}. HA: \textit{happy}. NE: \textit{neutral}. SA: \textit{sad}. SU: \textit{surprise}. CO: \textit{contempt}. AX: \textit{anxiety}. HL: \textit{helplessness}. DS: \textit{disappointment}.
	}
	\label{fig:confusion_matrix}
\end{figure*}

\subsubsection{Expert Activation Patterns Analysis}

{
	To investigate the task-adaptive behavior of the proposed MoAE module, we visualize the most frequently activated expert paths across layers, as shown in Fig. \ref{fig:expert_activation_paths}. Although the DFER and SFER tasks share the  FFN branch in the MoAE layer, the activated experts differ significantly across tasks. For instance, the expert selections in top-2 pathways for AffectNet-7 exhibit less overlap with those in FERV39K and DFEW, suggesting that \textbf{the gating mechanism effectively preserves task-specific token routing and facilitates the learning of task-dependent representations.} Notably, both DFER-related tasks (i.e., FERV39K and DFEW) and the SFER task (AffectNet-7) frequently activate expert E6 at layer L3, indicating the emergence of a shared mid-level expert. This observation implies that \textbf{the model captures partially task-agnostic representations, thus enabling structural coupling beyond FFN sharing and promoting knowledge transfer across related tasks.} Moreover, the activation paths in DFER exhibit greater consistency between the top-1 and top-2 selections, especially from layers L1 to L4, suggesting a more concentrated and deterministic routing pattern. In contrast, SFER demonstrates a more diverse expert selection strategy, relying on a larger number of expert activation pathways. These results indicate that MoAE simultaneously learns task-specific routing configurations while selectively reusing experts to encode cross-task shared features.  }

\subsection{Visualization}

\subsubsection{Visualization of Attentions}
Fig. \ref{fig:attention_map} presents the attention maps from both the baseline model and S4D across four emotion categories: \textit{sad}, \text{fear}, \textit{angry}, and \textit{surprise}.  The results demonstrate that integrating SFER knowledge enables our model to achieve more focused and precise attention compared to the baseline.
Specifically, for the \textit{sad} expression, the baseline model exhibits relatively dispersed attention, particularly around the mouth and forehead—regions that are generally less reliable for distinguishing sadness. In contrast, the S4D model effectively concentrates on the eyes, brows, and mouth, which are critical regions for recognizing sadness, as indicated by the activation of Action Units (AUs) 1 (Inner Brow Raiser), 4 (Brow Lowerer), and 15 (Lip Corner Depressor) \cite{ekman1978facial}. Similarly, for \textit{fear}, the baseline model primarily allocates attention to the upper face while neglecting key regions. Our model, however, effectively targets the eyes and mouth, which are essential for detecting fear, aligning with the activation of AUs such as 1 (Inner Brow Raiser), 5 (Upper Lid Raiser), and 26 (Jaw Drop) \cite{ekman1978facial}.  This trend is also evident in the attention maps for both \textit{angry} and \textit{surprise} expressions, where the S4D model demonstrates a more comprehensive and focused attention allocation than the baseline. In summary, incorporating SFER data significantly enhances the model’s attention to critical facial regions across all emotional categories, leading to a notable improvement in DFER performance over the baseline.

\subsubsection{Confusion Matrix Analysis}
We performed a comprehensive statistical analysis of S4D's performance on the DFEW~\cite{DFEW}, FERV39k~\cite{ferv39k2022}, and MAFW~\cite{MAFW} datasets, with the results presented through confusion matrices.
As depicted in Fig. \ref{fig:confusion_matrix}, majority categories such as \textit{happy}, \textit{sad}, and \textit{neutral} expressions demonstrate high accuracy across all datasets.  In contrast, minority categories, including \textit{disgust}, \textit{fear}, and \textit{helplessness} exhibit lower accuracy, particularly on the DFEW dataset. {This performance gap is likely attributable to the limited sample sizes and severe class imbalance in these categories. For example, disgust constitutes only 1.22\% of the total samples in DFEW.} Such challenges can be alleviated by implementing imbalanced learning strategies, such as oversampling techniques. Additionally, we observed that some expressions are frequently misclassified as \textit{neutral}, particularly \textit{disgust} and \textit{surprise}, \textbf{due to the subtle boundaries between \textit{neutral} and these emotional expressions.} For example, in both the DFEW and FERV39K datasets, \textit{disgust} is frequently misclassified as \textit{neutral}, likely because of the subtlety of \textit{disgust} expressions, such as slight nose wrinkling or minor lip movements.

\section{Conclusion}
In this paper, we proposed Static-for-Dynamic (S4D), a novel unified dual-modal learning framework that integrates SFER data to enhance DFER. By employing dual-modal pre-training and joint fine-tuning on both FER image and video datasets, S4D effectively learns powerful spatiotemporal representations, enabling a more comprehensive understanding of dynamic facial expressions and significantly advancing dynamic expression recognition. Furthermore, the proposed MoAE module, integrated into the latter ViT layers, empowers the model to better utilize learned generic representations for task-specific feature extraction. This design effectively mitigates negative transfer and promotes more discriminative feature learning tailored to DFER.
Additionally, despite the differences between SFER and DFER tasks, our analysis of the correlation between these tasks from both semantic and expert pathway perspectives highlights their complementary nature in dual-modal learning.
Extensive experimental evaluations on the FERV39K, MAFW, and DFEW benchmarks demonstrated the effectiveness and superiority of S4D, surpassing the baseline by large margins and setting a new state of the art.

However, achieving state-of-the-art performance for DFER requires a careful balance of SFER data during joint fine-tuning, which poses challenges in optimizing both DFER and SFER tasks simultaneously.  In future work, we will focus on developing more efficient architectures and optimization strategies to unify the learning of DFER, SFER, and other visual affective recognition tasks, aiming for comprehensive excellence across all aspects of these tasks.

\bibliographystyle{IEEEtran}
\bibliography{egbib}

\vspace{11pt}

\vspace{11pt}
\vfill

\end{document}